\documentclass[10pt,journal,compsoc]{IEEEtran}
\ifCLASSOPTIONcompsoc
  \usepackage[nocompress]{cite}
\else
  \usepackage{cite}
\fi
\ifCLASSINFOpdf
\else
\fi
\usepackage{graphicx}
\usepackage{comment}
\usepackage{amsmath,amssymb} 
\usepackage{color}
\usepackage[export]{adjustbox}
\usepackage{subcaption}
\usepackage[inline]{enumitem}
\usepackage{booktabs}
\usepackage{array,multirow,graphicx}
\usepackage{siunitx}
\usepackage{pifont}
\usepackage{textcomp}

\usepackage[implicit=false]{hyperref}
\hypersetup{
    colorlinks=true,
    urlcolor=blue,
}

\RequirePackage{color}
\definecolor{RED}{rgb}{1,0,0}
\definecolor{BLUE}{rgb}{0,0,1}
\definecolor{White}{rgb}{1,1,1}

\def\eg{\emph{e.g.,}} \def\Eg{\emph{E.g.}}
\def\ie{\emph{i.e.,}} 
 
\def\etc{\emph{etc.}}
\def\vs{\emph{vs. }}

\def\etal{\emph{et al.}}

\def\Vec#1{{\boldsymbol{#1}}}
\def\Mat#1{{\boldsymbol{#1}}}

\newcommand{\fig}{Fig.}
\newcommand{\tab}{Table}

\begin{document}
%
\title{Attention in Reasoning: Dataset, Analysis, and Modeling}
%
%
%
%

\author{Shi~Chen,
        Ming~Jiang,
        Jinhui~Yang,
        and~Qi~Zhao
\IEEEcompsocitemizethanks{\IEEEcompsocthanksitem
The authors are with the Department of Computer
Science and Engineering, University of Minnesota, Minneapolis, MN, 55455.
E-mail: see \url{http://www-users.cs.umn.edu/˜qzhao/}. The first two authors have equal contributions.}%
}

%
%

\markboth{Journal of \LaTeX\ Class Files,~Vol.~xx, No.~x, xx~xxxx}%
{Chen \MakeLowercase{\textit{et al.}}: ARC: Attention in the Reasoning Context}
%



\IEEEtitleabstractindextext{%
\begin{abstract}
While attention has been an increasingly popular component in deep neural networks to both interpret and boost the performance of models, little work has examined how attention progresses to accomplish a task and whether it is reasonable. In this work, we propose an Attention with Reasoning capability (AiR) framework that uses attention to understand and improve the process leading to task outcomes. We first define an evaluation metric based on a sequence of atomic reasoning operations, enabling a quantitative measurement of attention that considers the reasoning process. We then collect human eye-tracking and answer correctness data, and analyze various machine and human attention mechanisms on their reasoning capability and how they impact task performance. To improve the attention and reasoning ability of visual question answering models, we propose to supervise the learning of attention progressively along the reasoning process and to differentiate the correct and incorrect attention patterns. We demonstrate the effectiveness of the proposed framework in analyzing and modeling attention with better reasoning capability and task performance. The code and data are available at \url{https://github.com/szzexpoi/AiR}.
\end{abstract}

}

\maketitle

\IEEEdisplaynontitleabstractindextext

%
\IEEEpeerreviewmaketitle

\section{Introduction}
Recent progress in deep neural networks (DNNs) has resulted in models with significant performance gains in many tasks. Attention, as an information selection mechanism, has been widely used in various DNN models \cite{senet,ccnet,att_branch,mask_guide,updown,pyramid_sod,pyramid_sod,hierarchical_att,aoa_captioning,sca_captioning}, to improve their ability of localizing important parts of the inputs, as well as task performance. It also enables fine-grained analyses and understanding of the black-box DNN models, by highlighting important information in their decision-making process. Recent studies explored different machine attention mechanisms and showed varied degrees of agreement on where humans consider important in various vision tasks, such as image captioning~\cite{captioning_human_19,captioning_human_17} and visual question answering (VQA)~\cite{vqahat}.

\begin{figure*}
\centering
\includegraphics[width=1\linewidth]{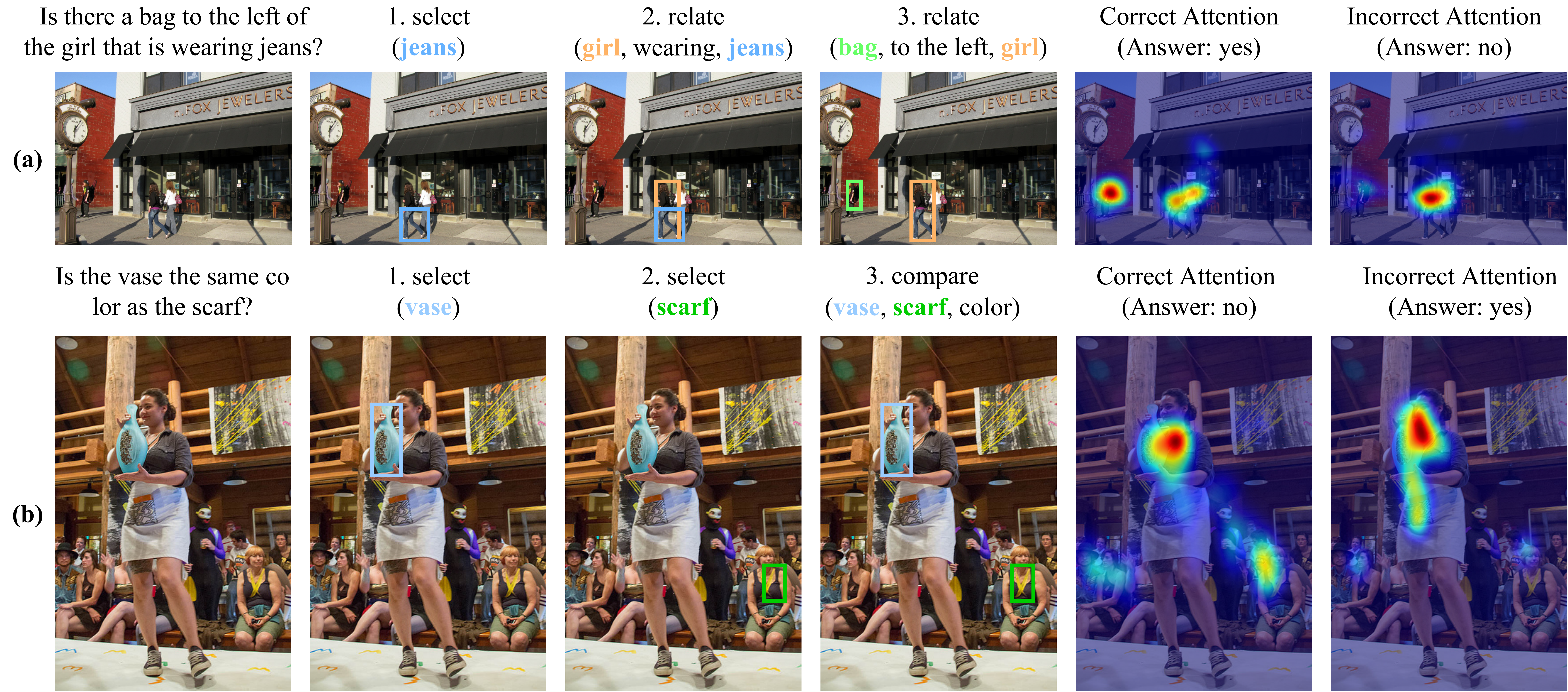}
\caption{Attention is an essential mechanism that affects task performances in visual question answering. (a) People who answer correctly look at the most relevant ROIs in the reasoning process (\ie~jeans, girl, and bag). (b) Incorrect answers can be caused by misdirected attention towards salient distractors (\ie~the skirt).}
\label{fig:teaser}
\end{figure*}

Similar to humans who look and reason actively and iteratively to perform a visual task, attention and reasoning are two intertwined mechanisms underlying the decision-making process. As shown in \fig~\ref{fig:teaser}, answering the question requires humans or machines to make a sequence of decisions based on the regions of interest (ROIs) (\ie~to sequentially look for the jeans, the girl wearing the jeans, and the bag to the left of the girl in \fig~\ref{fig:teaser}a), and avoid the distraction from visually salient but task-irrelevant information (\ie~the skirt in \fig~\ref{fig:teaser}b). Guiding attention to explicitly look for these objects following the reasoning process has the potential to improve both the interpretability and the performance of a computer vision model. 

To understand the role of visual attention in VQA, and leverage attention for model development, we propose an integrated Attention with Reasoning capability (AiR) framework. It represents the visual reasoning process as a sequence of atomic operations each with specific ROIs, defines a metric that enables the quantitative evaluation of attention, and proposes two supervision methods that guide attention based on the differentiation of attention patterns and the intermediate steps of the visual reasoning process. A new eye-tracking dataset is collected to support the understanding of human visual attention during the visual reasoning process and is also used as a baseline for studying machine attention. This framework is a useful toolkit for research in visual attention and its interaction with visual reasoning.

Our work has four distinctions from previous attention evaluation~\cite{vqahat,gqa,explanation,qual_eval} and supervision~\cite{paan,han,mining} methods: (1) We go beyond the existing evaluation methods that are either qualitative or focused only on the alignment of attention outputs, and propose a measure that encodes the progressive attention and reasoning defined by a set of atomic operations. (2) Focusing on the tight correlation among attention, reasoning, and task performance, we conduct fine-grained analyses to answer various research questions about different types of attention. (3) We jointly supervise machine attention with the reasoning data, so that it can progressively focus on different regions of interest in each step of the reasoning process. (4) We help machines avoid salient distractors by guiding their attention with both correct and incorrect attention patterns. (5) Our new dataset with human eye movements and answer correctness enables more accurate evaluation and diagnosis of attention.

To summarize, the proposed framework makes the following contributions:
\begin{enumerate}
    \item a new quantitative evaluation metric (AiR-E) to measure attention in the reasoning context, based on a set of constructed atomic reasoning operations,
    \item a progressive attention supervision method (AiR-M) to optimize the reasoning operations and the allocation of machine attention throughout the entire reasoning process,
    \item a correctness-aware attention supervision method (AiR-C) that for the first time incorporates both correct and incorrect attention to guide the learning of machine attention,
    \item an eye-tracking dataset (AiR-D) featuring high-quality attention and reasoning labels as well as ground truth answer correctness,
    \item extensive analyses of various human and machine attention mechanisms with respect to attention accuracy and task performance. Multiple factors of attention in the reasoning process have been examined and discussed. Experiments show the significance of the proposed attention dataset, evaluation metric, and supervision methods.
\end{enumerate}

In particular, this paper extends our preliminary study~\cite{air2020} in the following aspects:
\begin{enumerate}
    \item We propose a new AiR-C method that for the first time considers attention and answer correctness during the learning of attention. It jointly leverages both correct and incorrect attention patterns as positive and negative guidance to supervise  machine attention (Section~\ref{neg_supervision} and Section~\ref{neg_sup_sec}).
    \item We introduce a new hold-out test set for AiR-D to facilitate future research on attention modeling. It consists of new images, questions, answers, and eye-tracking data. The 406 questions of this new dataset are on a different set of 319 images. It provides a new benchmark for attention studies and can be used to test generalizability of models (Section~\ref{sec:benchmark}).
    \item We conduct new analyses about the inter-subject consistency of the eye-tracking data, and find that human attention in the VQA task is highly consistent (Section~\ref{sec:benchmark}).
    \item To demonstrate the impacts of task information on attention allocation, we conduct a new quantitative study to investigate the attention difference when answering different questions about the same image (Section~\ref{task_impact}).
    \item To understand how attention affects task performance, we explicitly compare the attention between correct and incorrect answers to the same question, which shows interesting observations and motivates the use of incorrect attention for models. (Section~\ref{diverse_att}).
    \item We conduct a new experiment to analyze the correlation as well as inconsistency between attention accuracy and reasoning performance, which suggests the significance of learning high-quality attention for visual reasoning (Section~\ref{att_ans_agreement}).
    \item We extend the ablation study for the proposed AiR-M method for more complete and thorough discussion (Section~\ref{att_supervision}). We have also explicitly discussed the advantages of the new AiR-C method with new evaluation results, in terms of improving the attention accuracy and answer accuracy (Section~\ref{neg_sup_sec}).
    \item We extend and reorganize Section~\ref{sec:related} to include a more comprehensive review of related studies. In particular, on human visual attention, we review attention datasets, models, and their applications in computer vision tasks (Section~\ref{sec:rw_hva}).
\end{enumerate}

The rest of the paper is organized as follows. In Section~\ref{sec:related}, we introduce the related studies on visual attention and reasoning. In Section~\ref{sec:framework}, we present the details of the proposed framework. Section~\ref{sec:analysis} reports the experiments and analyses on various attention mechanisms. Finally, in Section~\ref{sec:conclusion}, we conclude this paper and provide directions for future work.

\section{Related Works}\label{sec:related}
In this section, we briefly review related literature on human attention (Section~\ref{sec:rw_hva}),  evaluation of machine attention in VQA (Section~\ref{sec:rw_evqa}), supervision of machine attention in VQA (Section~\ref{sec:rw_svqa}), and visual reasoning datasets (Section~\ref{sec:rw_vrd}).

\begin{table}
\begin{center}
\scriptsize
\resizebox{0.49\textwidth}{!}{
\begin{tabular}{ccccc}
\toprule
   Dataset &  No. of Scenes & Task & HPA & RP\\
\midrule
MIT-1003~\cite{judd2009learning}& 1003 & PV & \ding{55} & \ding{55} \\
EMOd~\cite{fan2018emotional}  & 1019 & PV & \ding{55} & \ding{55}  \\
DHF1K~\cite{wang2019revisiting} & 1000 & PV & \ding{55} & \ding{55} \\
CAMO~\cite{video_saliency} & 120 & PV & \ding{55} & \ding{55} \\
Webpage Saliency~\cite{shen2014webpage} & 149 & Web browsing & \ding{55} & \ding{55} \\
EGTEA Gaze+~\cite{li2018eye} & 86 & Cooking & \ding{55} & \ding{55} \\
DR(eye)VE~\cite{palazzi2018predicting} & 74 & Driving & \ding{55} & \ding{55} \\
IQVA~\cite{iqva} & 975 & VQA & \ding{51} & \ding{55}\\
\midrule
AiR-D & 1828 & VQA & \ding{51} & \ding{51} \\
\bottomrule
\end{tabular}
}
\end{center}
\caption{A comparison between different eye-tracking datasets. PV: passive viewing. HPE: human performance annotation. RP: reasoning process. }\label{tab:stats}
\end{table}

\subsection{Human Visual Attention}
\label{sec:rw_hva}
This paper is related to a collection of human visual attention studies. Leveraging biologically-inspired filters, attention models compute a probability map that indicates where humans look when freely observing an image or a video~\cite{borji2012state}. Early computational models of attention focus on studying the bottom-up mechanism driven by visual stimuli. To evaluate attention models and train data-driven algorithms for attention prediction, many eye-tracking datasets~\cite{bruce2005saliency,cerf2008predicting,bovik2009doves,judd2009learning,ramanathan2010eye,kootstra2011predicting,xu2014predicting,jiang2014saliency,fan2018emotional,jiang2015salicon,borji2015cat2000,wang2019revisiting} have been developed. Unlike the bottom-up mechanism, the top-down mechanism directs human visual attention using a task, which attracts growing research interests. Eye-tracking datasets have been built to study where humans look in various vision tasks (see Table~\ref{tab:stats}), including visual search in 2D~\cite{ehinger2009modelling,gilani2015pet,zelinsky2019benchmarking,yang2020predicting} or 3D images~\cite{3D_saliency}, and dynamic videos~\cite{video_saliency,wang2019revisiting}, action recognition~\cite{mathe2014actions,lu2019deep}, web-browsing~\cite{shen2014webpage,zheng2018task}, cooking~\cite{li2018eye}, driving~\cite{palazzi2018predicting}, and video-gaming~\cite{zhang2020atari}. The above attention datasets have empowered data-driven models, especially deep neural networks with remarkable learning abilities~\cite{borji2019saliency}, so that the performance gap between attention prediction models and humans has been significantly reduced. Human attention datasets and models have also contributed to the development of many computer vision applications~\cite{att_survey_ijcv}, such as object recognition~\cite{salah2002selective,han2010biologically}, scene classification~\cite{borji2011scene}, salient object segmentation~\cite{wang2019inferring,wang2020paying}, video summarization~\cite{xu2015gaze},~\etc~In this work, to facilitate the analysis of human attention in VQA, we construct this new eye-tracking dataset collected from humans performing the VQA tasks.

\subsection{Evaluation of Machine Attention in VQA}
\label{sec:rw_evqa}
This paper is closely related to prior studies on the evaluation of machine attention mechanisms in VQA~\cite{vqahat,gqa,explanation,qual_eval}. In particular, the pioneering work by Das~\etal~\cite{vqahat} is the only one that collected human attention data for VQA and compared them with machine attention, showing considerable discrepancies in the attention maps. Our proposed study highlights several distinctions from related works:
\begin{enumerate*}
    \item[(1)] Instead of only considering one-step attention and its alignment with a single ground-truth map, we propose to integrate attention with progressive reasoning that involves a sequence of operations related to different objects.
    \item[(2)] While most VQA studies assume human answers to be accurate, it is not always the case~\cite{yang2018visual}. We collect ground truth correctness labels to examine the effects of attention and reasoning on task performance, and investigate how humans and machines prioritize their attention in search of diverse answers.
    \item[(3)] The only available dataset~\cite{vqahat}, with post-hoc attention annotation collected on blurry images using a ``bubble-like'' paradigm and crowdsourcing, may not accurately reflect the actual attention of the task performers~\cite{tracking_compare}. Our work addresses these limitations by using on-site eye-tracking data and QA annotations collected from the same participants.
    \item[(4)] Das~\etal~\cite{vqahat} only compared spatial attention with human attention. Since recent studies~\cite{gqa,qual_eval} suggest that attention based on object proposals is more semantically meaningful, we conduct the first quantitative and principled evaluation of object-based attention.
\end{enumerate*}

\subsection{Supervision of Machine Attention in VQA}
\label{sec:rw_svqa}
This paper presents supervision approaches for the learning of attention mechanisms for VQA, which is related to the recent efforts in improving machine attention accuracy with explicit supervision. Several studies use different sources of attention ground truth, such as human visual attention~\cite{han}, adversarial learning~\cite{paan}, and objects mined from textual descriptions~\cite{mining}, to explicitly supervise the learning of machine attention. Similar to the attention evaluation studies introduced above, these attention supervision studies only consider attention as a single-output mechanism, and optimize models to attend to all ROIs with a single glimpse. They typically lead to outspread attention maps that cannot focus on the most relevant regions. They are also agnostic to the reasoning process and fail to acquire sufficient information from intermediate reasoning steps. Besides, these methods only consider the attention ground truths that positively contribute to the correct answers, but do not explicitly identify salient distractors that may lead to incorrect answers. Our work addresses these challenges from two distinct perspectives: (1) jointly predicting the reasoning operations and the desired attention throughout the entire process, enabling the learning of progressive and reasoning-aware attention, and (2) supervising models with both correct and incorrect attention to improve their attention outputs and answers.

\begin{table*}[t]
\begin{center}
\resizebox{0.7\textwidth}{!}{
\begin{tabular}{cc}
\toprule
\textbf{Operation} & \textbf{Semantic}  \\
\midrule
Select & Searching for objects from a specific category. \\
\midrule
Filter & Determining the targeted objects by looking for a specific attribute. \\
\midrule
Query & Retrieving the value of a specific attribute from the ROIs. \\
\midrule
Verify & Examining the targeted objects and checking if they have a given attribute. \\
\midrule 
Compare & Comparing the values of an attribute between multiple objects. \\
\midrule
Relate & Connecting different objects through their relationships. \\
\midrule
And/Or & Serving as basic logical operations that combine the results of the previous operation(s). \\
\bottomrule
\end{tabular}
}
\end{center}
\caption{Semantic operations of the reasoning process.}
\label{semantic}
\end{table*}

\begin{figure*}
\centering
\includegraphics[width=0.6\linewidth]{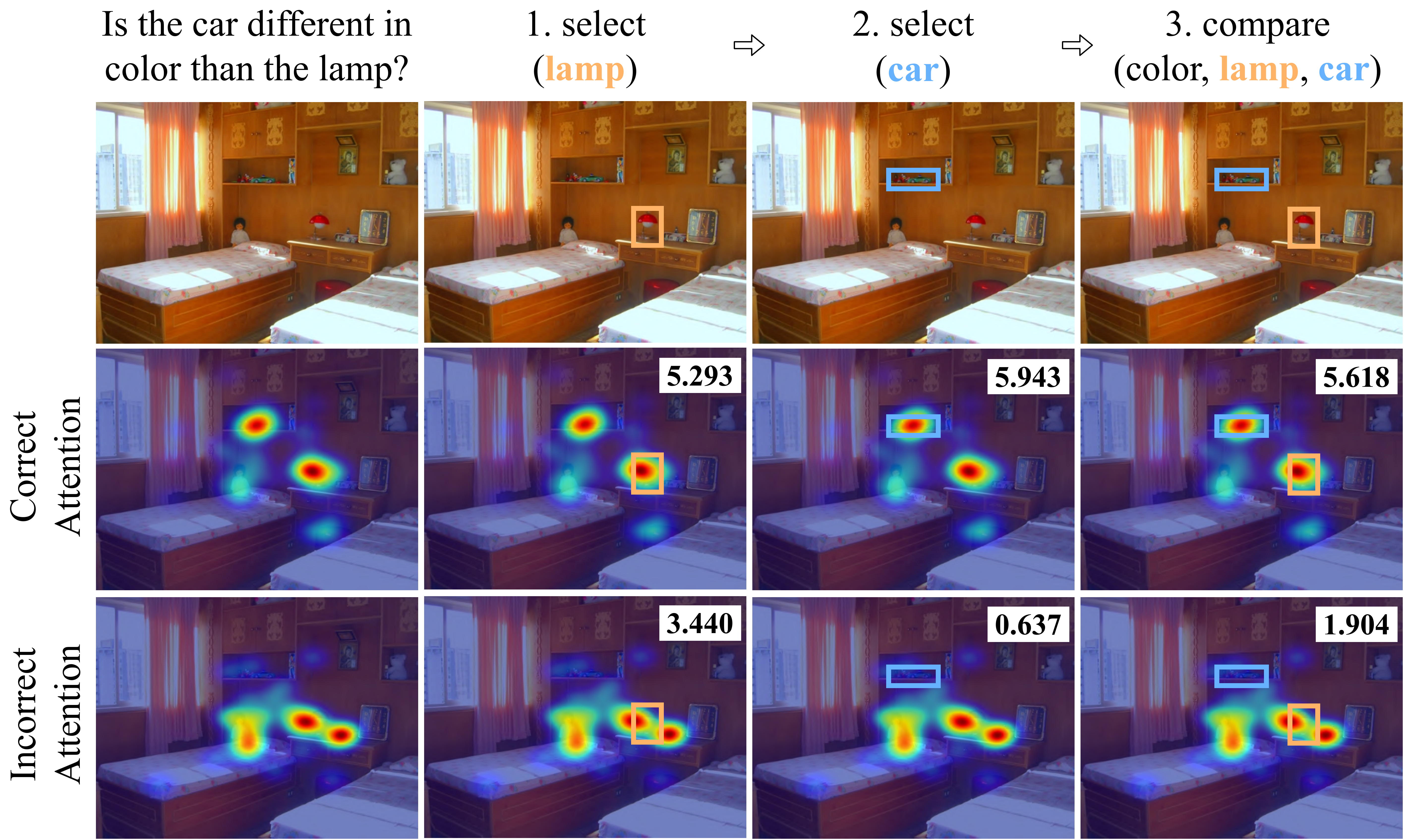}
\caption{AiR-E scores of Correct and Incorrect human attention maps. They measure the alignment between attention and the bounding boxes of ROIs.}
\label{fig:framework}
\end{figure*}

\subsection{Visual Reasoning Datasets}
\label{sec:rw_vrd}
Several visual reasoning datasets~\cite{vqa1,vqa2,gqa,clevr,movieqa,vcr,ok_vqa,TQA,est_vqa,knowit,tvqa} have been collected in the form of VQA. Some are annotated with human-generated questions and answers \cite{vqa1,movieqa}, while others are developed with synthetic scenes and rule-based templates to remove the subjectivity of human answers and the language bias~\cite{vqa2,gqa,clevr,vcr}. The one most closely related to this work is GQA~\cite{gqa}, which offers naturalistic images annotated with scene graphs and synthetic question-answer pairs. With balanced questions and answers, it reduces the language bias without compromising generality. Their data efforts benefit the development of various visual reasoning models~\cite{vqa_nips,updown,trilinear,mcb,n2nmn,ban,tbd-net,sc_vqa,san,ns_vqa,coatt,mfb,hint,counter_vqa,regat}. In this work, we use a selection of GQA data and annotations in the development of the proposed framework.

\section{Method}\label{sec:framework}
Real-life vision tasks require looking and reasoning interactively. This section presents a principled framework to study attention in the reasoning context. It consists of three novel components:
\begin{enumerate*}
  \item[(1)] a quantitative measure to evaluate attention accuracy for visual reasoning,
  \item[(2)] a progressive supervision method for models to learn where to look throughout the reasoning process, and
  \item[(3)] an eye-tracking dataset featuring human eye-tracking and answer correctness data.
\end{enumerate*}

\subsection{Attention with Reasoning Capability} \label{reasoning_progress}

To model attention as a process and examine its reasoning capability, we describe reasoning as a sequence of atomic operations. Following the sequence, an intelligent agent progressively attends to the key ROIs at each step and reasons what to do next until eventually making a final decision. A successful decision-making method relies on accurate attention for various reasoning operations, so that the most important information is not filtered out but passed throughout to the final step.

To represent the reasoning process and obtain the corresponding ROIs, we define a vocabulary of atomic operations emphasizing the role of attention. These operations are grounded on the 127 types of operations of GQA \cite{gqa} that completely represent all questions. We define the operations by characterizing and abstracting the complex functional programs of the GQA dataset. Specifically, we define each operation as a triplet, \ie~$<$operation, attribute, category$>$, and categorize the original operations in the GQA program based on their semantic meanings:
\begin{enumerate*}
    \item[(1)] For the original operations that exactly align with our definitions, we directly convert them into our triplet representation, for example, from ``filter size table'' to $<$filter, large/small, table$>$; 
    \item[(2)] If the original operations do not have an exact match, we convert them into our operations with similar semantic meanings. For example, we convert ``different color object A and object B'' to $<$compare, color, category A and category B$>$.
\end{enumerate*} 
As described in \tab~\ref{semantic}, our operations cover various situations for attention allocation: some require attention to a specific object (\textit{query}, \textit{verify}); some require attention to objects of the same category (\textit{select}), attribute (\textit{filter}), or relationship (\textit{relate}); and others require attention to any (\textit{or}) or all (\textit{and}, \textit{compare}) ROIs from the previous operations.

\begin{figure*}
\centering
\includegraphics[width=0.8\linewidth]{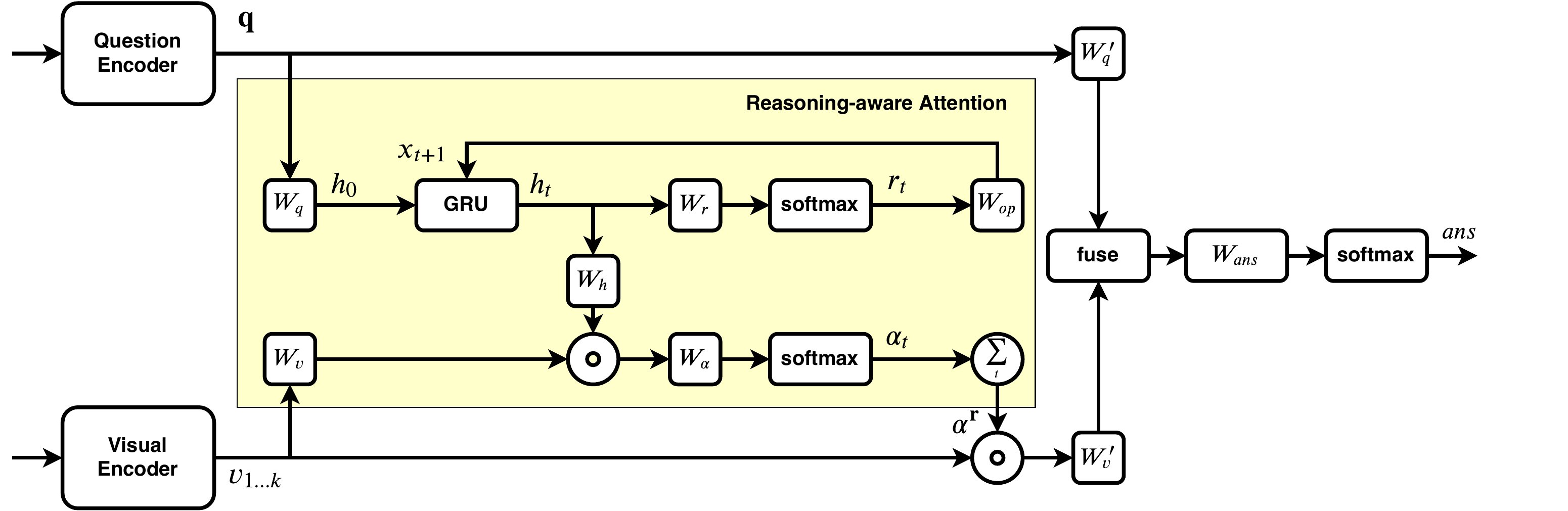}
\caption{Network architecture of the proposed AiR-M method.}
\label{fig:model}
\end{figure*}

Upon obtaining the operation for each reasoning step, we determine its corresponding ROIs by jointly considering both the semantic meaning of operation and the scene information (\ie~object categories, attributes, and relationships in the scene graphs):
\begin{itemize}
    \item \textbf{Select}: The ROIs belong to a specific category of objects. We query all objects in the scene graph and select those with the same category as defined in the triplet.
    \item \textbf{Query, Verify}: The ROIs are defined in a similar way as the ``select'' operation. The difference is that they are selected from the ROIs of the previous step, instead of the entire scene graph.
    \item \textbf{Filter}: The ROIs are a subset of the previous step's ROIs with the same attribute as defined in the triplet.
    \item \textbf{Compare, And, Or}: These operations are based on multiple groups of objects. Therefore, the ROIs are the combination of all ROIs of the related previous reasoning steps.
    \item \textbf{Relate}: The ROIs are a combination of two groups of objects: the ROIs of the previous reasoning step and a specific category of objects from the scene graph.
\end{itemize}

Some questions in GQA~\cite{gqa}, \eg~``Is there a red bottle on top of the table'' with answer ``no'', refer to non-existing objects. In such cases, we select the $k$ most frequently co-existent objects as the ROIs. Specifically, based on the scene graphs, we first compute the frequency of co-existence between different object categories on the training set. Next, given a particular reasoning operation referring to a nonexistent object, the top-$k$ ($k=20$) co-existing objects in the scene are selected as the corresponding ROIs.

The aforementioned paradigm allows us to efficiently traverse the reasoning process, starting with all objects in the scene, and sequentially investigating the operation and ROIs at each step. It enables the evaluation of attention accuracy throughout the continuous reasoning process (Section~\ref{sec:AiR-E}), and the progressive supervision that guides models to learn where to look following the process (Section~\ref{sup_method}).

\subsection{Measuring Attention Accuracy with ROIs}\label{sec:AiR-E}
Decomposing the reasoning process into a sequence of operations allows us to evaluate the quality of machine or human attention according to its alignment with the ROIs at each operation. Attention can be represented as a 2D probability map where values indicate the importance of the corresponding input pixels. To quantitatively evaluate attention accuracy in the reasoning context, we propose the AiR-E metric that measures the alignment of the attention maps with ROIs relevant to reasoning. As shown in \fig~\ref{fig:framework}, for humans, a better attention map leading to the correct answer has higher AiR-E scores, while the incorrect attention with lower scores fails to focus on the most important object (\ie~car). It suggests a potential correlation between the AiR-E and the task performance. The specific definition of AiR-E is introduced as follows:

Inspired by the Normalized Scanpath Saliency \cite{sal_metric} (NSS), given an attention map $A(x)$ where each value represents the importance of a pixel $x$, we first standardize the attention map into $A^*(x) = \left(A(x) - \mu\right)/\sigma$,
where $\mu$ and $\sigma$ are the mean and standard deviation of the attention values in $A(x)$, respectively. For each ROI, we compute AiR-E as the average of $A^*(x)$ inside its bounding box $B$: $\text{AiR-E}(B) =  \sum\limits_{x \in B} A^*(x)/|B|$. Finally, we aggregate the AiR-E of all ROIs for each reasoning step:
\begin{enumerate}
    \item For operations with one set of ROIs (\ie~\textit{select}, \textit{query}, \textit{verify}, and \textit{filter}) \textit{or} that requires attention to one of the multiple sets of ROIs, an accurate attention map should align well with at least one ROI. Therefore, the aggregated AiR-E score is the maximum AiR-E of all ROIs.
    \item For those with multiple sets of ROIs (\ie~\textit{relate}, \textit{compare}, \textit{and}), we compute the aggregated AiR-E for each set and take the mean across all sets.
\end{enumerate}

\subsection{Reasoning-aware Attention Supervision} \label{sup_method}
For models to learn where to look throughout the reasoning process, we propose a reasoning-aware attention supervision method (AiR-M) to guide models to progressively look at relevant places following each reasoning operation. Different from previous attention supervision methods~\cite{paan,han,mining}, the AiR-M method considers the attention throughout the reasoning process and jointly supervises the prediction of reasoning operations and ROIs across the sequence of multiple reasoning steps. Integrating attention with reasoning allows models to accurately capture ROIs throughout the entire reasoning process for deriving the correct answers.

As shown in \fig~\ref{fig:model}, the proposed method has two major distinctions:
\begin{enumerate*}
    \item[(1)] integrating attention progressively throughout the entire reasoning process and 
    \item[(2)] joint supervision on attention, reasoning operations, and answer correctness.
\end{enumerate*}

Specifically, following the reasoning decomposition discussed in Section~\ref{reasoning_progress}, the proposed method takes the question features $q$ and the visual features $V$ as the inputs, and uses a Gated Recurrent Unit \cite{gru} (GRU) to sequentially predict the operations $\Vec{r}_t$ and the desired attention weights $\Vec{\alpha}_t$ at the $t$-th step. At the beginning of the reasoning process, the hidden state of GRU $\Vec{h}_{0}$ with the question features $\Vec{q}$ is defined as:
\begin{equation}
    \Vec{h}_{0} = \Mat{W}_{q} \Vec{q},
\end{equation}
\noindent where $\Mat{W}_{q}$ represents trainable weights. We update the hidden state $\Vec{h}_{t}$, and simultaneously predict the reasoning operation $\Vec{r}_{t}$ and attention $\Vec{\alpha}_t$:
\begin{equation}
    \Vec{r}_{t} = \text{softmax} (\Mat{W}_{r} \Vec{h}_{t}),
\end{equation}
\begin{equation}\label{att_pred}
    \Vec{\alpha}_t = \text{softmax}( \Mat{W}_{\alpha} (\Mat{W}_{v} \Vec{v} \circ \Mat{W}_{h} \Vec{h}_{t}) )
\end{equation}
\noindent where $\Mat{W}_{r}, \Mat{W}_{\alpha}, \Mat{W}_{h}$ are all trainable weights, and $\circ$ is the Hadamard product. The next step input $\Vec{x}_{t+1}$ is computed with the predicted operation:
\begin{equation}
    \Vec{x}_{t+1} = \Mat{W}_{op} \Vec{r}_{t}
\end{equation}
\noindent where $\Mat{W}_{op}$ represents the weights of an embedding layer. 
By iterating over the whole sequence of reasoning steps, we compute the aggregated reasoning-aware attention
\begin{equation}
  \Vec{\alpha}^r =  \sum\limits_{t} \Vec{\alpha}_t / T
\end{equation}
that takes into account all intermediate attention weights along the reasoning process, where $T$ is the total number of reasoning steps. With the supervision from the ROIs for different reasoning steps, the model can adaptively aggregate attention over time to perform complex visual reasoning.

With the joint prediction of the operation $\Vec{r}_t$ and the attention $\Vec{\alpha}_t$, models learn desirable attention for capturing the ROIs throughout the reasoning process and deriving the answer. The predicted operations and attention outputs are supervised together with the prediction of answers:
\begin{equation}
    L = L_{ans} + \theta \sum\limits_{t} L_{\Vec{\alpha_t}} +  \phi \sum\limits_{t}  L_{\Vec{r}_t}
\end{equation}
\noindent where $\theta$ and $\phi$ are hyperparameters. We use the standard cross-entropy loss $L_{ans}$ and $L_{\Vec{r}_t}$ to supervise the answer and operation prediction, and a Kullback–Leibler divergence loss $L_{\Vec{\alpha}_t}$ to supervise the attention prediction. We aggregate the loss for operation and attention predictions over all reasoning steps.

The ground-truth attention map $L_{\Vec{\alpha_t}}$ is derived from our decomposed reasoning process. Specifically, we first extract ROIs for the current reasoning step $t$, and then compute the Intersection of Union (IoU) between each ROI and each input region proposal~\cite{updown}. The attention weight for each region proposal is defined as the sum of its IoUs with all ROIs. Finally, the ground truth attention map is constructed by normalizing the attention weights with the sum of weights for all input region proposals.

The proposed AiR-M supervision method is general and can be applied to various models with attention mechanisms. In the supplementary materials, we illustrate the implementation details for integrating AiR-M with different state-of-the-art models used in our experiments.

\subsection{Correctness-aware Attention Supervision} \label{neg_supervision}
Successful visual reasoning requires not only attention to regions of interest throughout the decision-making process, but also avoiding visually salient distractors that commonly lead to problematic answers. To address this need, we further propose a Correctness-aware Attention Supervision method (AiR-C) that uses both correct and incorrect attention patterns to guide the learning of machine attention. The differentiation between the correct and incorrect attention patterns reveals important cues for visual reasoning: on the one hand, correct attention captures the ROIs most relevant to the task, providing essential information that leads to the correct answer. On the other hand, the incorrect attention highlights the salient distractors that commonly lead to wrong answers (Section \ref{diverse_att}), and enables the models to avoid these hard-negative regions. To our best knowledge, despite many efforts on the learning of attention \cite{han,mining,paan}, we are the first to propose the usage of incorrect attention in the supervision of machine attention.

Specifically, we introduce supervision of the incorrect attention using a negative cross-entropy loss:
\begin{equation}
L_{att}^{-} =  \sum\limits_{p} M^{-}_{p} \log \alpha_{p}   
\end{equation}
\noindent where $M^{-}$ denotes the incorrect attention map, $\alpha$ represents the predicted model attention, and $p$ corresponds to different positions within the maps. The loss encourages models to avoid the distractors, while at the same time allows them to freely explore the other positions. The overall training objective can be formulated as follows:
\begin{equation}
    L = L_{ans} + \theta L_{att}^{+} +  \phi L_{att}^{-}
\end{equation}
\noindent where $L_{ans}$ is the answer prediction loss, $L_{att}^{+}$ is the attention loss for correct attention (\textit{e.g.}, \cite{mining}), $\theta$ and $\phi$ are the hyperparameters. 

Given a question and the corresponding image, we construct the ground truth incorrect attention maps by mining the top-$k$ frequently mentioned ROIs in other questions on the same image. We empirically set $k=3$ in our experiments and exclude those highly overlapping with the relevant ROIs. The rest are considered as hard-negative ROIs used to supervise the incorrect attention. The overlapping area between two ROIs is measured as the proportion of intersection $I$:
\begin{equation}
    I_{j,k} = \frac{O_{j}^{-} \cap O_{k}^{+}}{min(O_{j}^{-},O_{k}^{+})}
\end{equation}
\noindent where $O^{-}$ denotes the mined ROIs and $O^{+}$ represents the ROIs relevant to the question. For the $k_{th}$ mined ROI $O_{k}^{-}$, we iteratively compute its overlapping areas with every ROI in $O^{+}$. If the maximum area is smaller than a threshold (\ie~0.3), we consider $O_{k}^{-}$ as a valid hard-negative ROI. The aforementioned method efficiently locates the hard-negative ROIs that are visually salient, but irrelevant to the given question. Finally, the selected hard-negative ROIs are aggregated into an incorrect attention map to guide the training of models.

\subsection{Evaluation Benchmark and Human Attention Baseline}
\label{sec:benchmark}
Previous attention data collected under passive image viewing~\cite{mit_sal}, approximations with post-hoc mouse clicks~\cite{vqahat}, or visually grounded answers~\cite{explanation} may not accurately or completely reflect human attention in the reasoning process. They also do not explicitly verify the correctness of human answers. To demonstrate the effectiveness of the proposed evaluation metric and supervision method, and to provide a benchmark for attention evaluation, we construct the first eye-tracking dataset for VQA. It, for the first time, enables the step-by-step comparison of how humans and machines allocate attention during visual reasoning.

Specifically, we 
\begin{enumerate*}
\item[(1)] select images and questions that require humans to actively look and reason;
\item[(2)] remove ambiguous or ill-formed questions and verify the ground truth answer to be correct and unique;
\item[(3)] collect eye-tracking data and answers from the same human participants, and evaluate their correctness with the ground-truth answers.
\end{enumerate*}

\textbf{Images and questions.} Our images and questions are selected from the balanced validation set of GQA~\cite{gqa}. Since the questions of the GQA dataset are automatically generated from a number of templates based on scene graphs~\cite{krishna2017visual}, the quality of these automatically-generated questions may not be sufficiently high. Some questions may be too trivial or too ambiguous. Therefore, we perform automated and manual screenings to control the quality of the questions. First, to avoid trivial questions, all images and questions are screened with these criteria:
\begin{enumerate*}
\item[(1)] image resolution is at least 320$\times$320 pixels;
\item[(2)] image scene graph consists of at least 16 relationships;
\item[(3)] total area of question-related objects does not exceed 4\% of the image.
\end{enumerate*}
Next, one of the authors manually selects 987 images and 1,422 questions to ensure that the ground-truth answers are accurate and unique. The selected questions are non-trivial and free of ambiguity, which requires paying close attention to the scene and actively searching for the answer.

In addition, to facilitate future research on task-driven attention modeling, we also introduce a new hold-out test set that contains 319 images and 406 questions. 
The average answer accuracy of the questions is 65.42\%, with a 26.38\% standard deviation. Eye-tracking data on this test set will not be released to the public. This test set will provide a new benchmark for gaze prediction in the visual reasoning context and will enable studies on the generalizability of attention modeling across different questions and answers.

\textbf{Eye-tracking experiment.} The eye-tracking data are collected from 20 paid participants, including 16 males and 4 females from age 18 to 38. They are asked to wear a Vive~Pro~Eye headset with an integrated eye-tracker to answer questions from images presented in a customized Unity interface. The questions are randomly grouped into 18 blocks, each shown in a 20-minute session. The eye-tracker is calibrated at the beginning of each session. During each trial, a question is first presented, and the participant is given unlimited time to read and understand it. The participant presses a controller button to start viewing the image. The image is presented in the center for 3 seconds. The image is scaled such that both the height and width occupy 30 degrees of visual angle (DVA). After that, the question is shown again and the participant is instructed to provide an answer. The answer is then recorded by the experimenter. The participant presses another button to proceed to the next trial.
\begin{figure}
    \centering
    \subfloat[]{\includegraphics[width=0.6\linewidth]{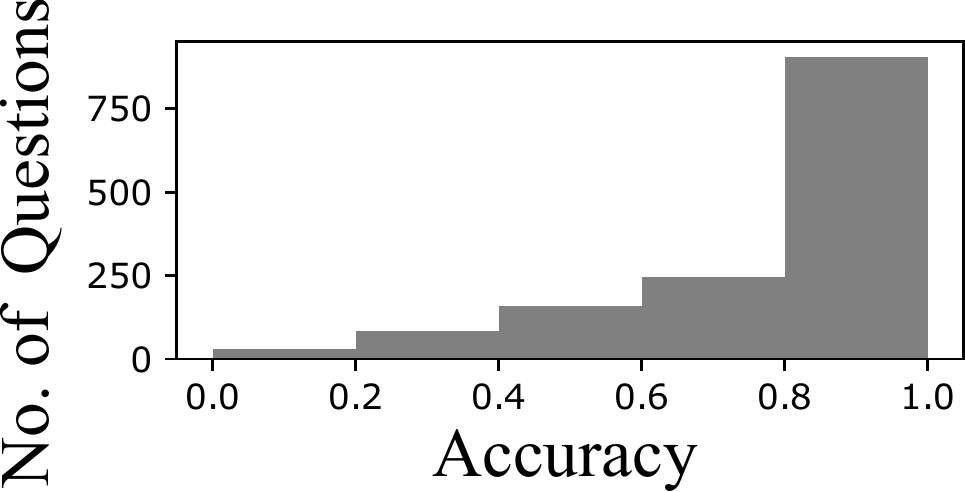}}
    \qquad
    \subfloat[]{\includegraphics[width=0.6\linewidth]{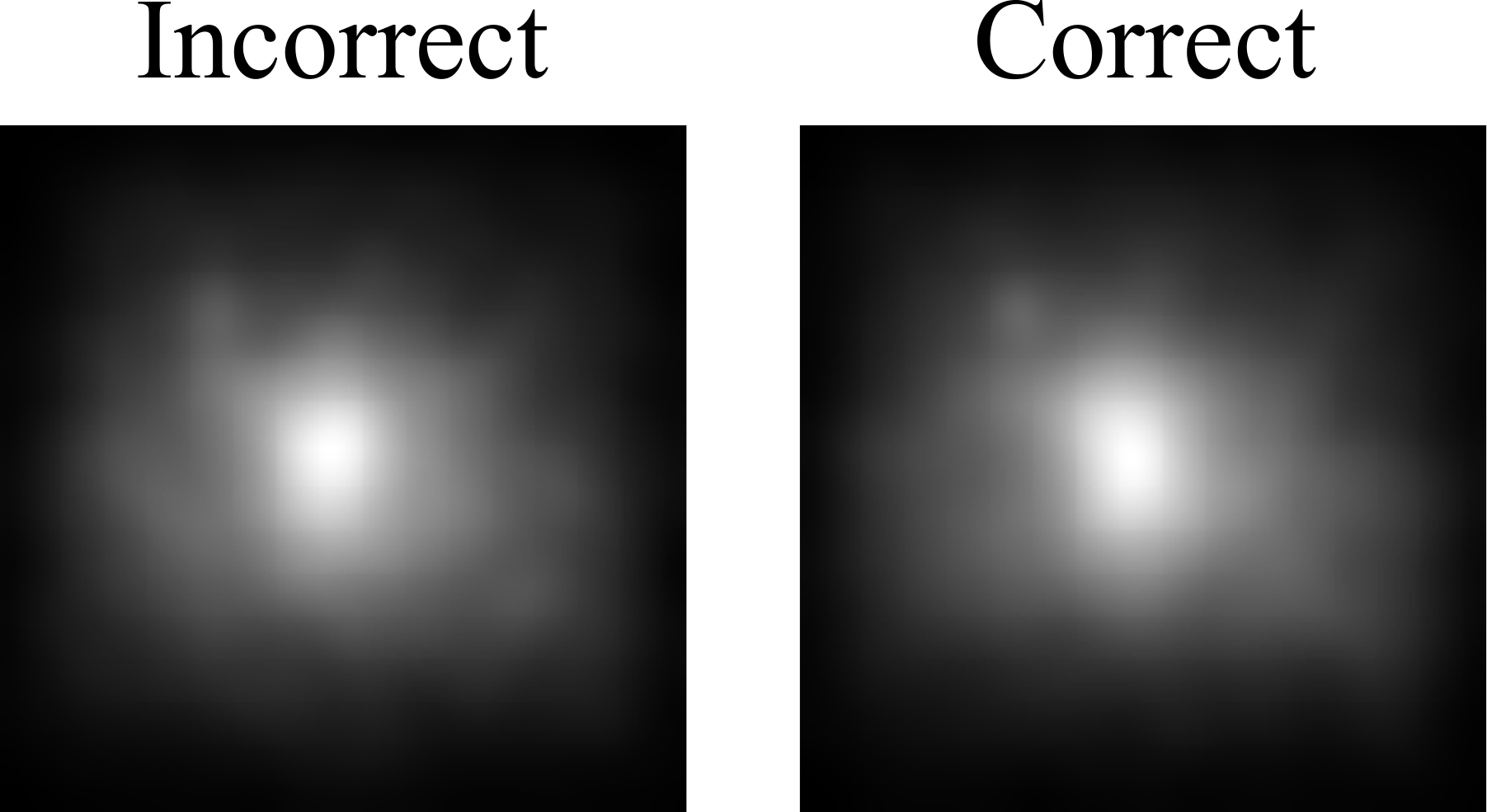}}
    \caption{Distributions of answer accuracy and eye fixations of humans. (a) Histogram of human answer accuracy (b) Center biases of the correct and incorrect attention.}
    \label{fig:et_stats}
\end{figure}

\textbf{Human attention maps and performances.} Eye fixations are extracted from the raw data using the Cluster Fix algorithm~\cite{cluster_fix}, and a fixation map is computed for each question by aggregating the fixations from all participants. The fixation maps are scaled into $256\times 256$ pixels, smoothed using a Gaussian kernel ($\sigma=9$ pixels, $\approx 1$ DVA), and normalized to the range of [0,1]. The overall accuracy of human answers is $77.64\pm 24.55\%$ (M$\pm$SD). A total of 479 questions have consistently correct answers, and 934 have both correct and incorrect answers. The histogram of human answer accuracy is shown in 
\fig~\ref{fig:et_stats}a. 
To quantify the inter-subject consistency in eye fixations, following~\cite{li2014secrets}, we randomly select data from half of the subjects and evaluate their fixation maps against the other half using the AUC-Judd~\cite{bylinskii2015saliency} metric. We observe a high inter-subject consistency (\ie~AUC-Judd=0.895) of eye fixations in the VQA task, which suggests the existence of consistently important visual cues that attract human attention in order to answer the questions.
We further separate the fixations into two groups based on answer correctness and compute a fixation map for each group. Correct and incorrect answers have comparable numbers of fixations per trial (10.12 \vs 10.27), while the number of fixations for the correct answers has a lower standard deviation across trials (0.99 \vs 1.54). \fig~\ref{fig:et_stats}b shows the prior distributions of the two groups of fixations, and their high similarity (Pearson's $r=0.997$) suggests that the answer correctness is independent of center bias. The correct and incorrect fixation maps are considered as two human attention baselines to compare with machine attention outputs, and also play a role in validating the effectiveness of the proposed AiR-E metric. Illustrative
examples are presented in the supplementary video.

\begin{figure*}
\centering
\includegraphics[width=0.9\linewidth]{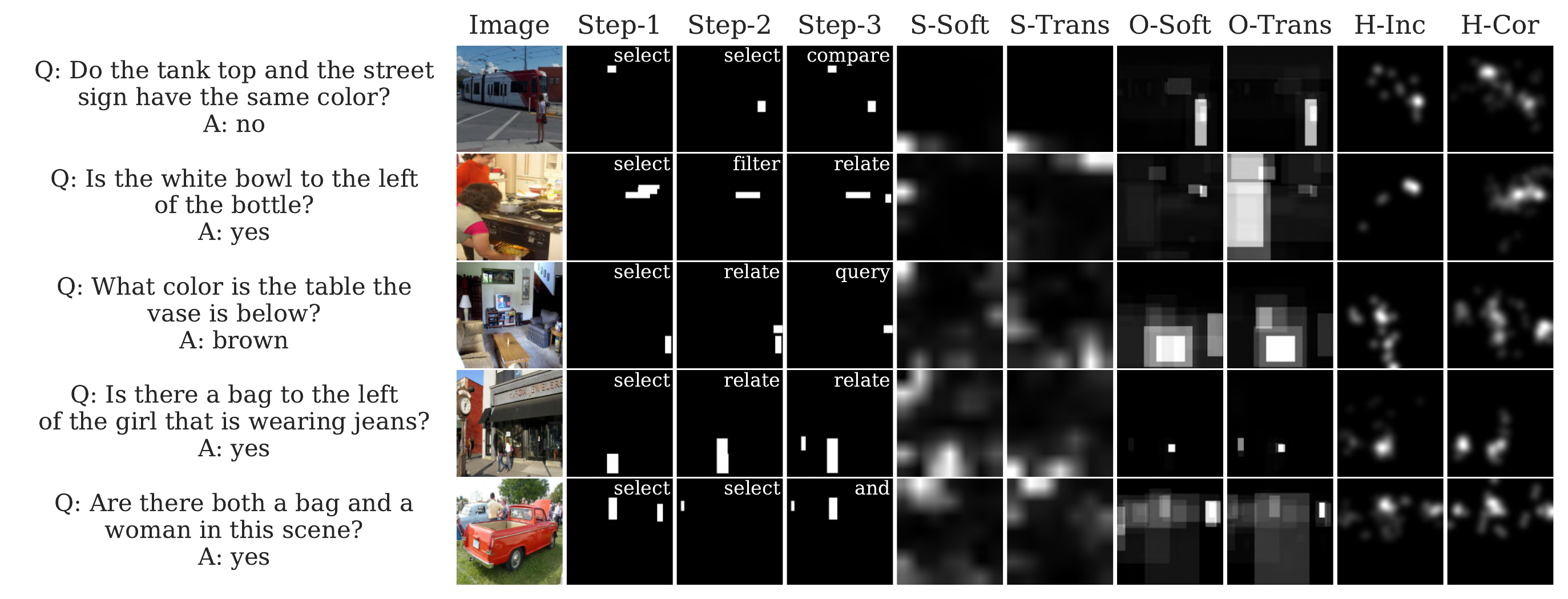}
\caption{Example question-answer pairs (column 1), images (column 2), ROIs at each reasoning step (columns 3-5), and attention maps (columns 6-11).}
\label{fig:qual}
\end{figure*}

\section{Experiments and Analyses}\label{sec:analysis}

In this section, we conduct experiments and analyze various attention mechanisms of humans and machines. 
Our experiments aim to shed light on the following questions that have yet to be answered:
\begin{enumerate}
    \item How do questions affect human attention? (Section~\ref{task_impact})
    \item Do machines or humans look at places relevant to the reasoning process? How does the attention process influence task performances? (Section~\ref{att_outcome})  
    \item How does attention accuracy evolve, and what about its correlation with the reasoning process? (Section~\ref{att_process})
    \item Do machines and humans with diverse answers look differently? (Section~\ref{diverse_att})
    \item Does progressive attention supervision improve attention and task performance? (Section~\ref{att_supervision}) 
    \item Is incorporating the incorrect attention beneficial for attention learning? (Section~\ref{neg_sup_sec})
    \item Do attention accuracy and reasoning performance agree? (Section \ref{att_ans_agreement})
\end{enumerate}

\subsection{How Do Questions Affect Human Attention?} \label{task_impact}
Human attention is driven by both the bottom-up visual stimuli and the top-down task information (\textit{e.g.}, question in the VQA task). 
Our AiR-D dataset has 299 images with at least two corresponding questions.  
To study how questions affect human attention in the VQA task, on this subset of eye-tracking data, we study the agreement between the fixation maps when answering two different questions about the same image.

Our experiments consider three distinct aspects of human attention (\ie~temporal dynamics, spatial alignment, and semantic alignment). For temporal dynamics, we group the fixations into three temporal bins (0-1s, 1-2s, and 2-3s), and compare the fixation maps for each bin. For spatial alignment, we compare fixation maps using Spearman's Rank correlation $r$ to measure the similarity of fixation distributions. For semantic alignment, we measure the average attention value in each object category of the scene, compare the top-5 object categories with the highest attention values, and evaluate the proportion of overlapping categories with Intersection over Union (IoU).

\begin{table}
\begin{center}
\begin{tabular}{ccccccc}
\toprule
& Spatial & Semantic \\
\midrule
Aggregated & 0.709 & 0.605  \\
\midrule
0-1s & 0.678 & 0.675 \\
1-2s & 0.536 & 0.590\\
2-3s & 0.439 & 0.564\\ 
\bottomrule
\end{tabular}
\end{center}
\caption{Spatial and semantic alignment scores between aggregated attention and attention over time.}
\label{alignment}
\end{table}

Table \ref{alignment} shows the spatial and semantic alignments of attention between different questions and their temporal evolutions. 
Two key observations can be drawn from the results: (1) There exists a considerable overlap between human attentions when answering different questions on the same image. This is validated by the relatively high spatial and semantic alignment scores (\ie~0.6) of their overall attention. (2) Both the spatial and semantic alignments decrease monotonically over time, suggesting that the question information progressively affects attention. At the beginning of visual exploration, people answering different questions focus on similar regions to quickly and broadly understand the image. After that, they gradually shift their attention towards the ROIs specific to each question, which results in low alignments between their attention patterns.

These observations show that human eye fixations have generally strong agreements when looking at the same image, even when answering different questions. However, the question information affects human attention in a dynamic manner, as the spatial and semantic agreements between attention patterns in different questions decrease monotonically over time.

\subsection{Do Machines or Humans Look at Places Important to Reasoning? How Does Attention Influence Task Performances?} \label{att_outcome}
In this subsection, we measure the attention accuracy throughout the reasoning process with the proposed AiR-E metric. Answer correctness is also compared, and its correlation with attention accuracy reveals the joint influence of attention and reasoning operations on task performance. With these experiments, we observe that humans attend more accurately than machines, and the correlation between attention accuracy and task performance depends on the reasoning operations.

\setlength{\tabcolsep}{0.45em}
\begin{table*}[t]
\begin{center}
\begin{tabular}{clllllllll}
\toprule
& Attention &             and &         compare &           filter &              or &           query &           relate &           select &          verify \\
\midrule
\parbox[t]{2mm}{\multirow{7}{*}{\rotatebox[origin=c]{90}{AiR-E}}} &H-Tot     &  2.197 &    2.669 &   2.810 & 2.429 &  3.951 &   3.516 &   2.913 &   3.629 \\
&H-Cor     &  2.258 &    2.717 &   2.925 & 2.529 &  4.169 &   3.581 &   2.954 &   3.580 \\
&H-Inc     &  1.542 &    1.856 &   1.763 & 1.363 &  2.032 &   2.380 &   1.980 &   2.512 \\
\cmidrule{2-10}
&O-Soft     &  1.334 &    \textbf{1.204} &   \textbf{1.518} & 1.857 & \textbf{3.241} &   \textbf{2.243} &   \textbf{1.586} &   2.091 \\
&O-Trans     &  \textbf{1.579} &    1.046 &   1.202 & \textbf{1.910} &  3.041 &   1.839 &   1.324 &   \textbf{2.228} \\
&S-Soft     & -0.001 &   -0.110 &   0.251 & 0.413 &  0.725 &   0.305 &   0.145 &   0.136 \\
&S-Trans     &  0.060 &   -0.172 &   0.243 & 0.343 &  0.718 &   0.370 &   0.173 &   0.101 \\
\midrule
\parbox[t]{2mm}{\multirow{5}{*}{\rotatebox[origin=c]{90}{Accuracy}}}&H-Tot     & 0.700 &    0.625 &   0.668 & 0.732 &  0.633 &   0.672 &   0.670 &   0.707 \\
\cmidrule{2-10}
&O-Soft     & 0.604 &    \textbf{0.547} &   0.603 & 0.809 &  \textbf{0.287} &   0.483 &   0.548 &   0.605 \\
&O-Trans     & \textbf{0.606} &    0.536 &   \textbf{0.608} & \textbf{0.832} &  0.282 &   \textbf{0.487} &   \textbf{0.550} &   0.592 \\
&S-Soft     & 0.592 &    0.520 &   0.558 & 0.814 &  0.203 &   0.427 &   0.511 &   0.544 \\
&S-Trans     & 0.597 &    0.525 &   0.557 & 0.811 &  0.211 &   0.435 &   0.517 &   \textbf{0.607} \\

\bottomrule
\end{tabular}
\end{center}
\caption{Quantitative evaluation of AiR-E scores and task performance.}
\label{AiR-E}
\end{table*} 

\begin{table*}
\begin{center}
\begin{tabular}{lllllllll}
\toprule
Attention &             and &         compare &           filter &              or &           query &           relate &           select &          verify \\
\midrule
H-Tot     &       0.205 &  \textbf{0.329} &         0.051 &  0.176 &  \textbf{0.282} &   \textbf{0.210} &   \textbf{0.134} &  \textbf{0.270} \\
\midrule
O-Soft     &       0.167 &  \textbf{0.217} &        -0.022 &  0.059 &  \textbf{0.331} &         0.058 &         0.003 &        0.121 \\
O-Trans     &       0.168 &  \textbf{0.205} &         0.090 &  0.174 &  \textbf{0.298} &         0.041 &   \textbf{0.063} &       -0.027 \\
S-Soft     &       0.177 &  \textbf{0.237} &        -0.084 &  0.082 &       -0.017 &  -0.170 &  -0.084 &        0.066 \\
S-Trans     &       0.171 &  \textbf{0.210} &  -0.152 &  0.086 &       -0.024 &  -0.139 &  -0.100 &  \textbf{0.270} \\
\bottomrule
\end{tabular}
\end{center}
\caption{Pearson's $r$ between attention accuracy (AiR-E) and task performance. Bold numbers indicate significant positive correlations (p$<$0.05).}
\label{corr}
\end{table*} 

We evaluate four types of attention that are commonly used in VQA models, including spatial soft attention (\ie~S-Soft), spatial Transformer attention (\ie~S-Trans), object-based soft attention (\ie~O-Soft), and object-based Transformer attention (\ie~O-Trans). Spatial and object-based attention differ in terms of their inputs (\ie~image features or regional features), while soft and Transformer attention methods differ in terms of the computational methods of attention (\ie~with convolutional layers or matrix multiplication). We use spatial features extracted from ResNet-101 \cite{resnet} and object-based features from~\cite{updown} as the two types of inputs, and follow the implementations of \cite{updown} and \cite{dfaf} for the soft attention~\cite{soft_att} and Transformer attention~\cite{mul_att} computation, respectively. We integrate the aforementioned attention mechanisms with different state-of-the-art VQA models as backbones. Our observations are general and consistent across various backbones. In the following sections, we use the results on UpDown~\cite{updown} for illustration (results for the other backbones are provided in the supplementary materials). For human attention, we denote the fixation maps associated with correct and incorrect answers as H-Cor and H-Inc, and the overall fixation map regardless of correctness is denoted as H-Tot. \fig~\ref{fig:qual} presents examples of ROIs for different reasoning operations and the compared attention maps. 

\textbf{Attention accuracy and task performance of humans and models.} \tab~\ref{AiR-E} quantitatively compares the AiR-E scores and VQA task performance across humans and models with different types of attention. The task performance for models is the classification score of the correct answer, while the task performance for humans is the proportion of correct answers. Three clear gaps can be observed from the table:
\begin{enumerate*}
\item[(1)] Humans who answer correctly have significantly higher AiR-E scores than those who answer incorrectly.
\item[(2)] Humans consistently outperform models in both attention and task performance. 
\item[(3)] Object-based attention mechanisms attend much more accurately than spatial attention.
\end{enumerate*}
The low AiR-E of spatial attention confirms the previous conclusion drawn from the VQA-HAT dataset~\cite{vqahat}. By constraining the visual inputs to a set of semantically meaningful objects, object-based attention typically increases the probabilities of attending to the correct ROIs. Between the two object-based attention, the soft attention slightly outperforms its Transformer counterpart. Since the Transformer attention explicitly learns the inter-object relationships, they perform better for logical operations (\ie~\textit{and}, \textit{or}). However, due to the complexity of the scenes and fewer parameters used~\cite{mul_att}, they do not perform as well as soft attention.
The ranks of different attention mechanisms are consistent with the intuition and literature, suggesting the effectiveness of the proposed AiR-E metric.

\begin{figure*}
\centering
    \subfloat[]{\includegraphics[width=0.32\linewidth]{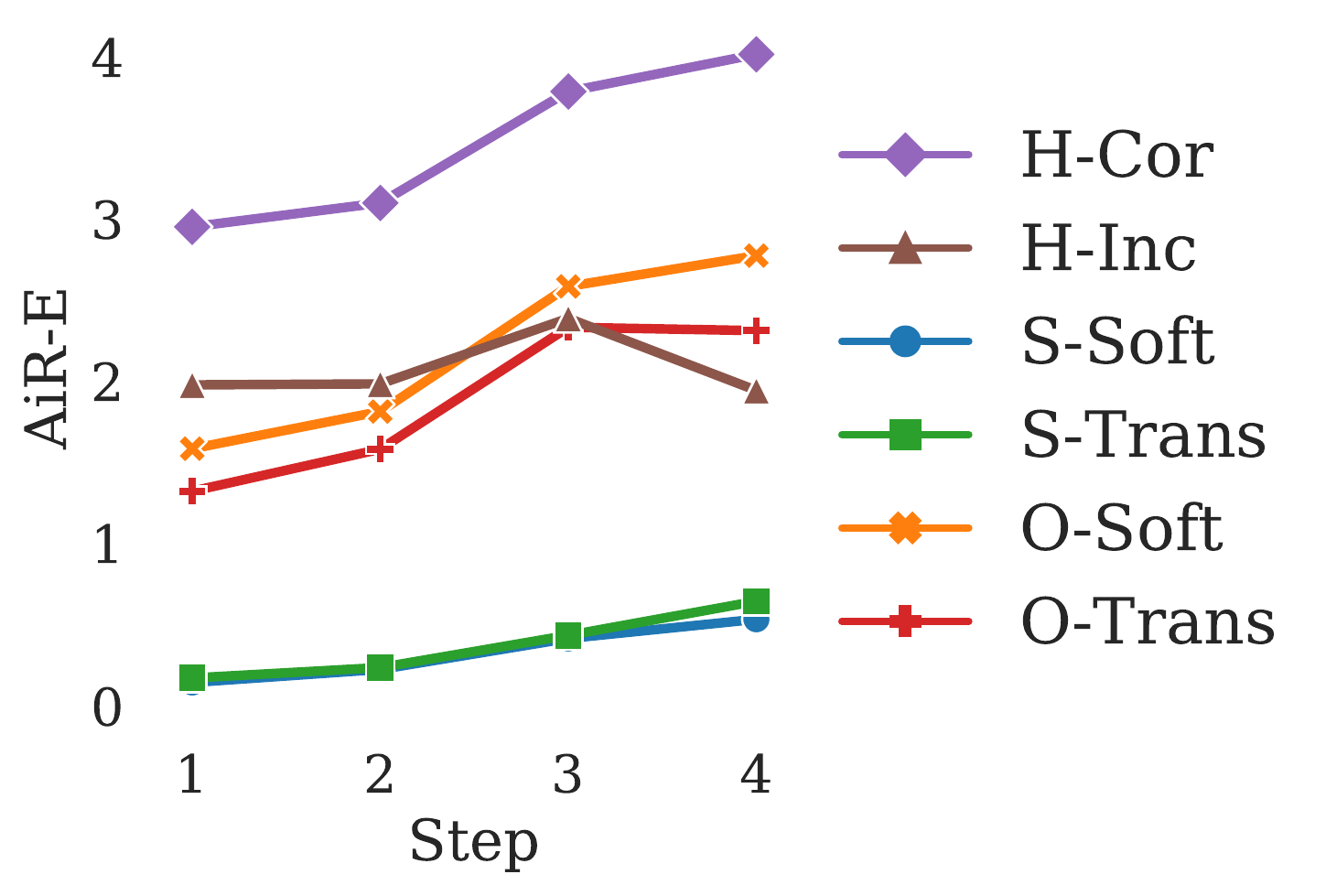}}
     \vrule
     \hfill
    \subfloat[]{\includegraphics[width=0.275\linewidth]{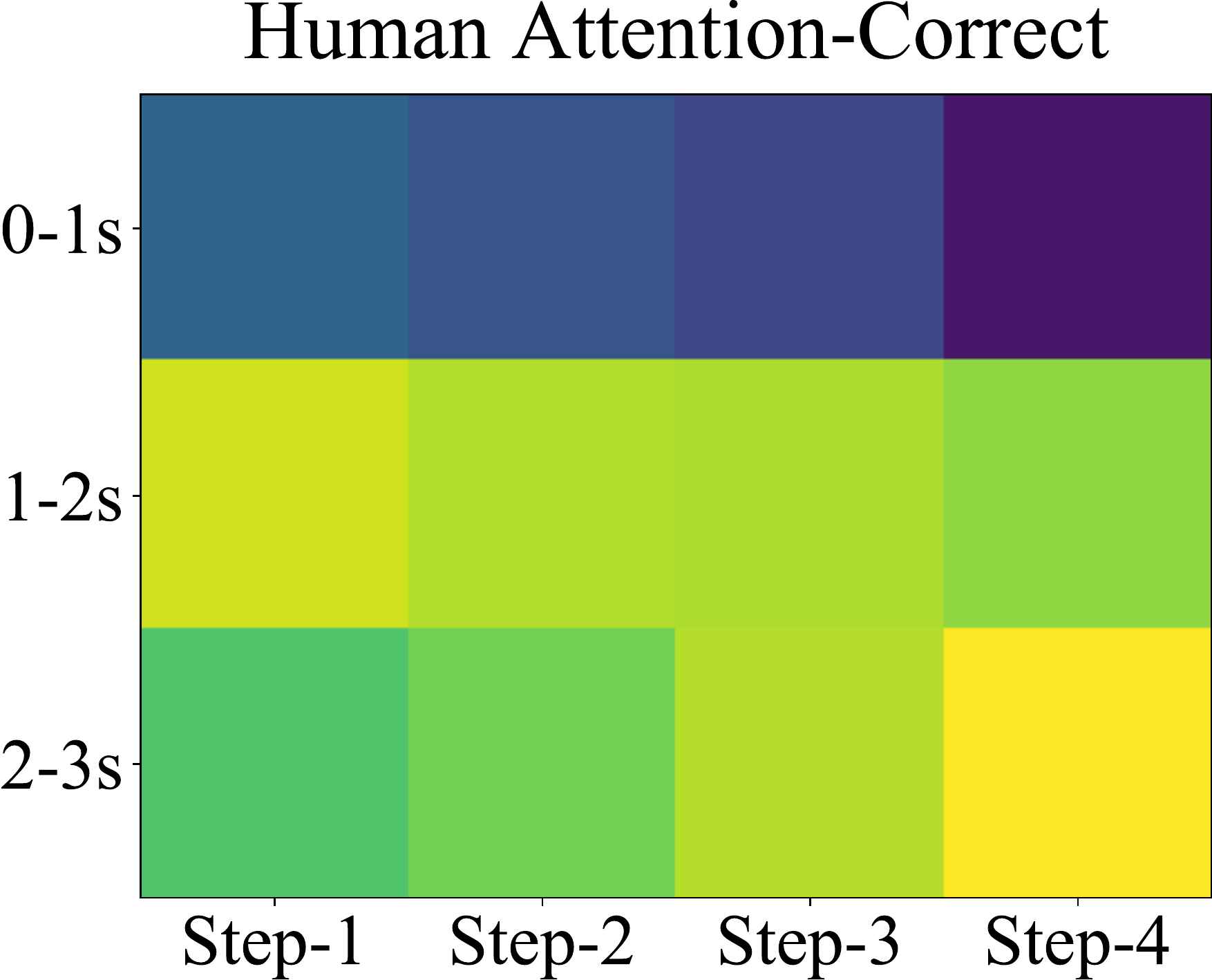}}
    \hfill
    \subfloat[]{\includegraphics[width=0.32\linewidth]{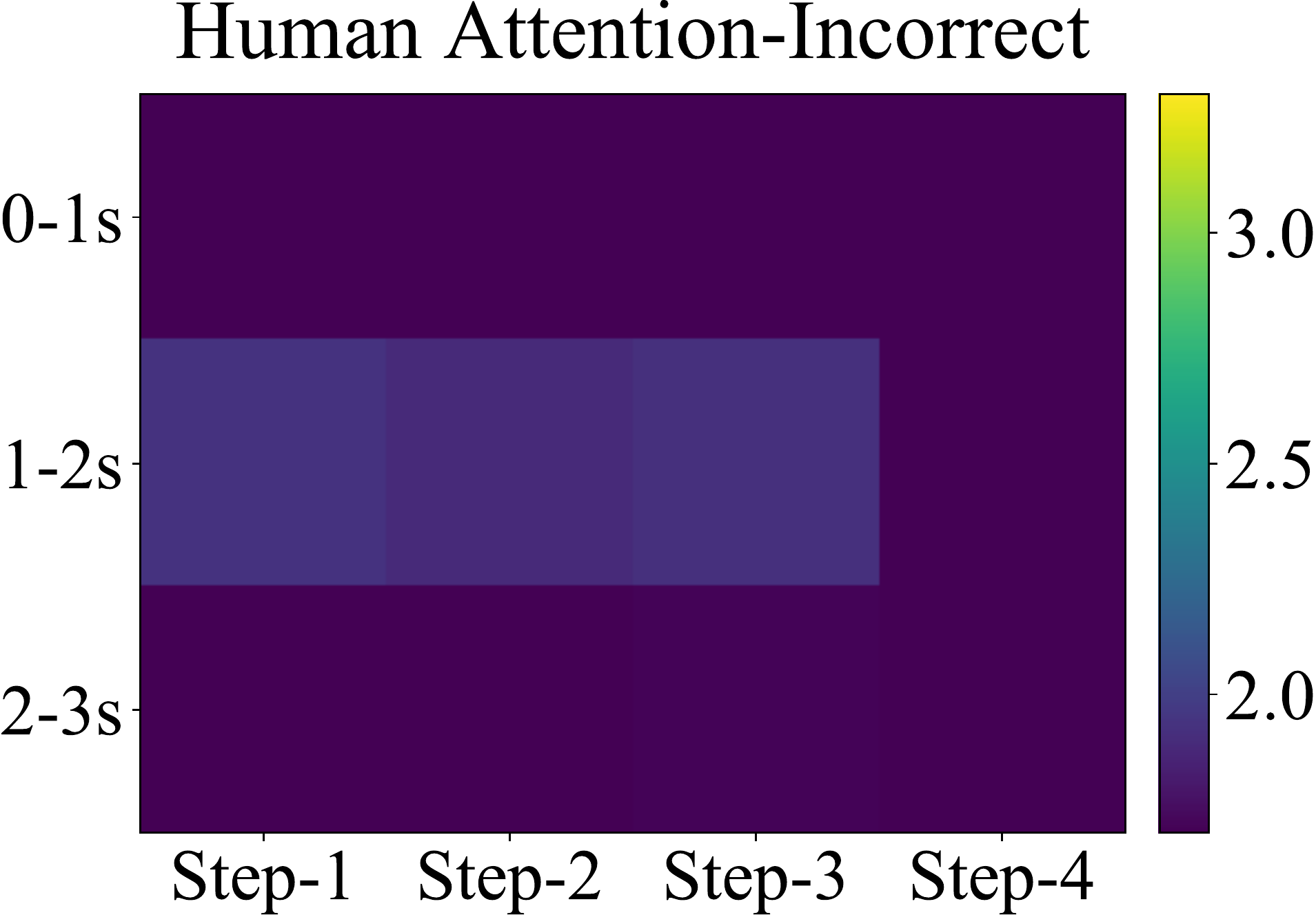}}
    \hfill
    \\
    \vspace{5pt}
    \hrulefill\\
    \vspace{5pt}
    \subfloat[]{\includegraphics[width=0.275\linewidth]{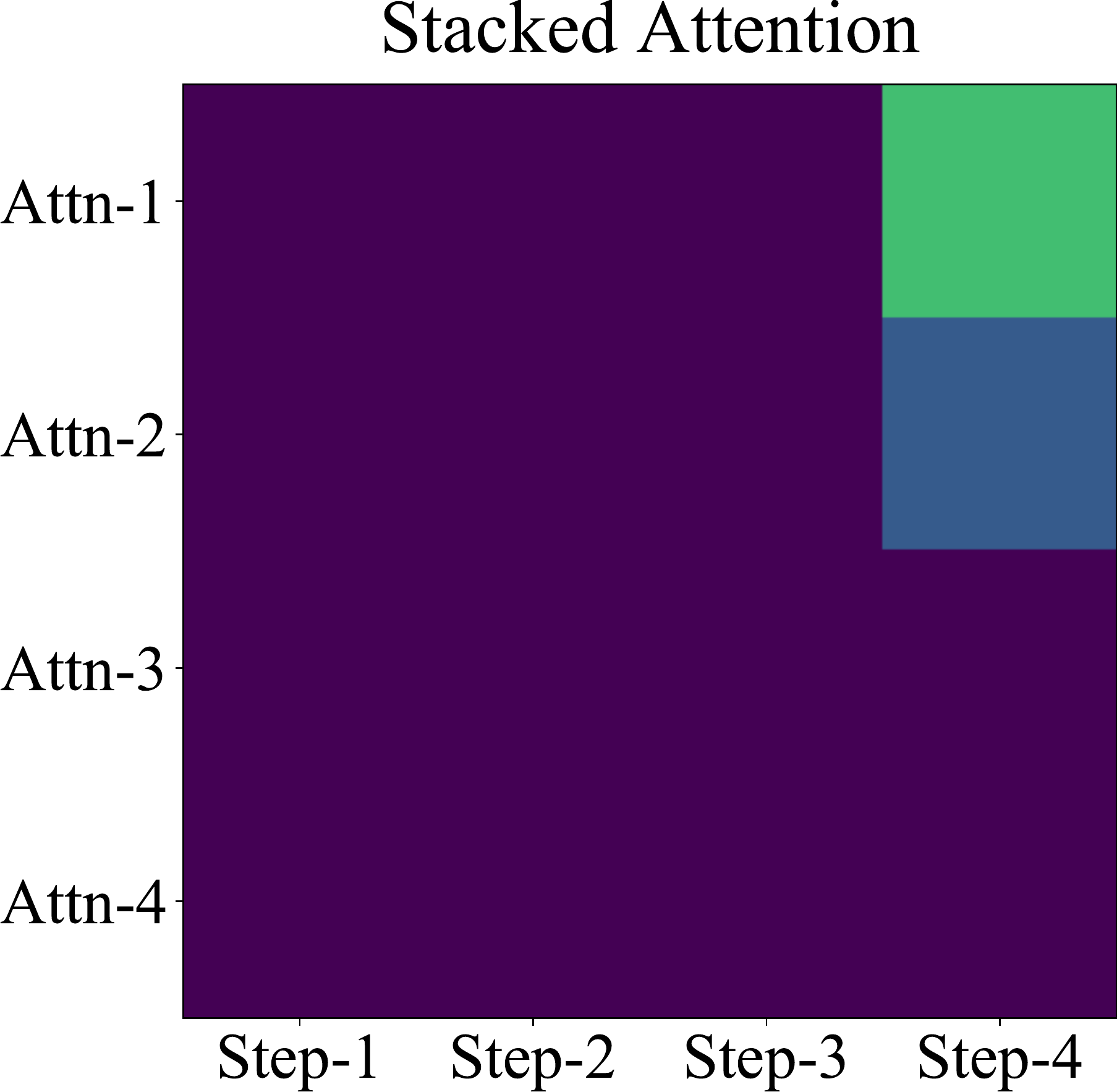}}
    \hfill
    \subfloat[]{\includegraphics[width=0.275\linewidth]{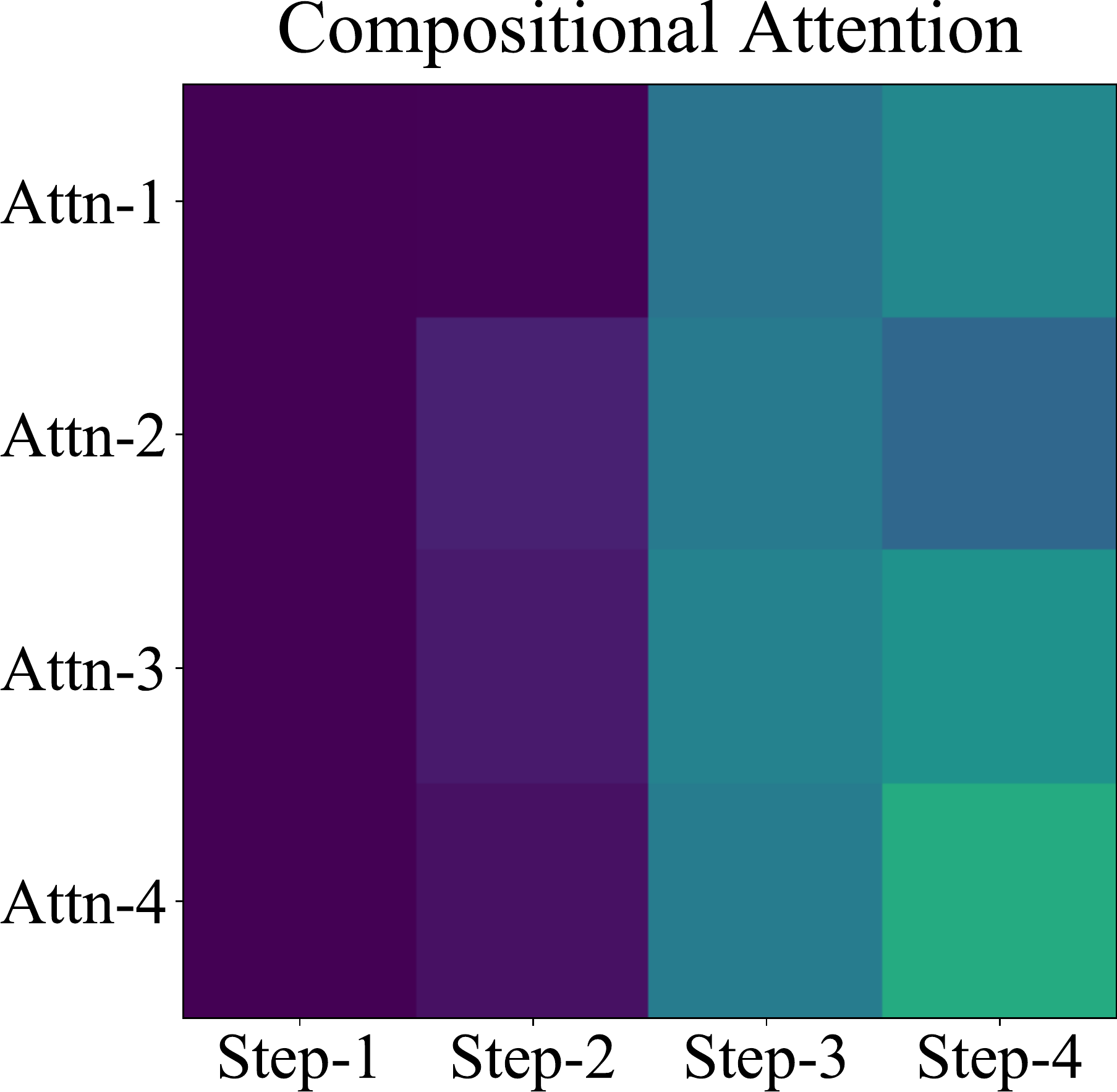}}
    \hfill
    \subfloat[]{\includegraphics[width=0.32\linewidth]{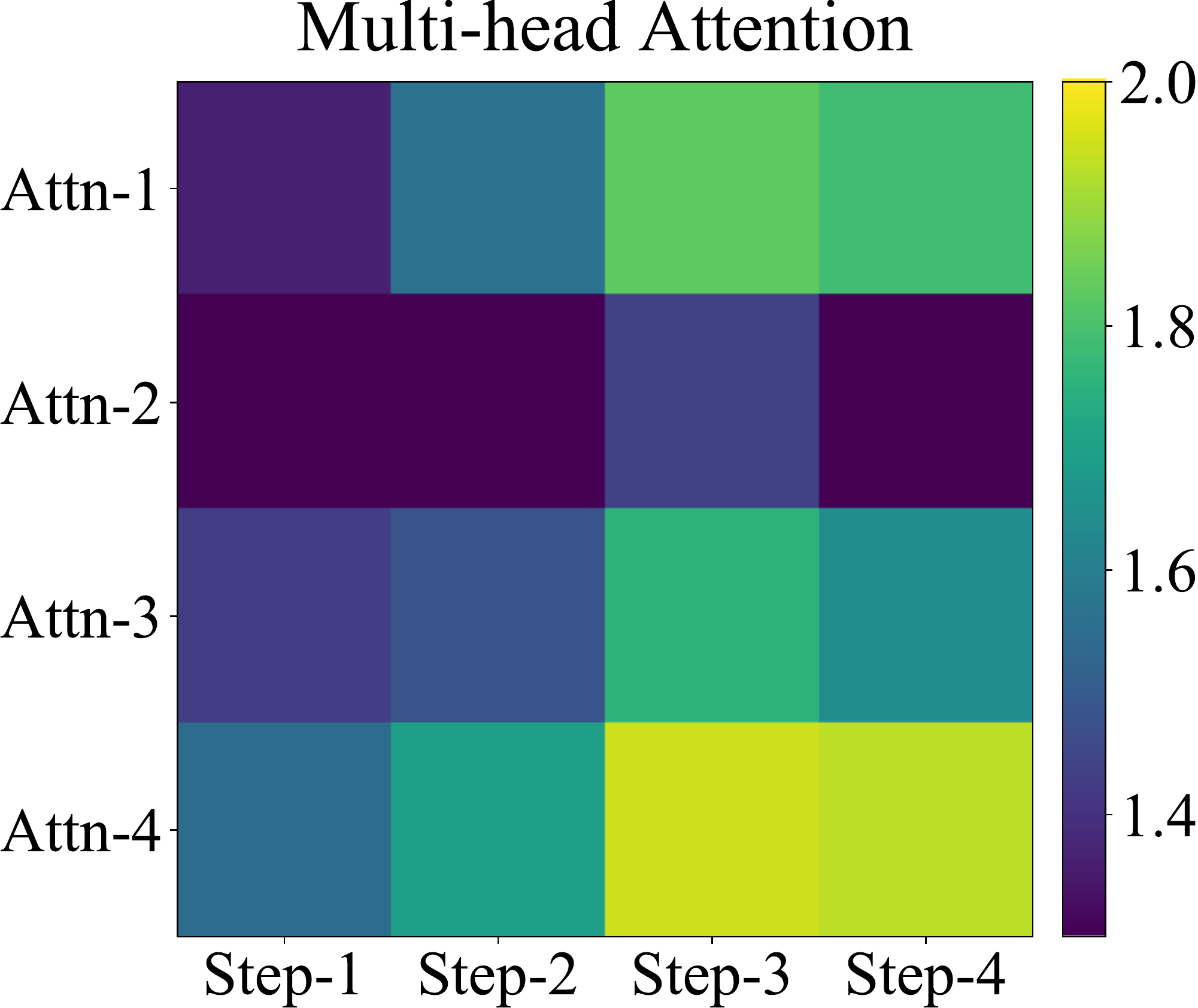}}       
  
  \caption{Spatiotemporal accuracy of attention throughout the reasoning process. (a) shows the AiR-E of different reasoning steps for human aggregated attention and single-glimpse machine attention, (b)-(c) AiR-E scores for decomposed human attention with correct and incorrect answers, (d)-(f) AiR-E for multi-glimpse machine attention. For heat maps shown in (b)-(f), the x-axis denotes the different reasoning steps while the y-axis corresponds to the indices of attention maps.} 
  \label{fig:step_comparison}
\end{figure*}

\textbf{Attention accuracy and task performance among different reasoning operations.} Comparing the different operations, \tab~\ref{AiR-E} shows that \textit{query} is the most challenging operation for models. Even with the highest attention accuracy among all operations, the task performance is the lowest. This is probably due to the inferior recognition ability of the models compared with humans. To humans, `compare' is the most challenging in terms of task performance, largely because it often appears in complex questions that require close attention to multiple objects and thus take longer processing time. Since models can process multiple input objects in parallel, their performance is not highly influenced by the number of objects to look at.

\textbf{Correlation between attention accuracy and task performance.} The similar rankings of AiR-E and task performance suggest a correlation between attention accuracy and task performance. To further investigate this correlation on a sample basis, for each attention and operation, we compute Pearson's $r$ between the attention accuracy and task performance across different questions.

As shown in \tab~\ref{corr}, human attention accuracy and task performance are correlated for most of the operations (up to $r=0.329$). The correlation is higher than most of the compared machine attention mechanisms, suggesting that humans' task performance is more consistent with their attention quality. In contrast, though commonly referred to as an interface for interpreting models' decisions~\cite{vqahat,explanation,qual_eval}, spatial attention maps do not reflect the decision-making process of models. They typically have very low and even negative correlations (\eg~\textit{relate}, \textit{select}). By limiting the visual inputs to foreground objects, object-based attention mechanisms achieve higher attention-answer correlations.

The differences in correlations between operations are also significant. For questions requiring focused attention to answer (\ie~with \textit{query} and \textit{compare} operations), the correlations are relatively higher than the others. 

\subsection{How Does Attention Accuracy Evolve Throughout the Reasoning Process?} \label{att_process}
To complement our previous analysis on the spatial allocation of attention, we move forward to analyze the spatiotemporal alignment of attention. Specifically, we analyze the AiR-E scores according to the chronological order of reasoning operations. We show in \fig~\ref{fig:step_comparison}a that the AiR-E scores peak at the 3rd or 4th steps, suggesting that human attention and machine attention focus more on the ROIs closely related to the final task outcome, instead of the earlier steps. In the rest of this section, we focus our analysis on the spatiotemporal alignment between multiple attention maps and the ROIs at different reasoning steps. In particular, we study the change of human attention over time and compare it with multi-glimpse machine attention. Our analysis reveals the significant spatiotemporal discrepancy between human attention and machine attention.

\textbf{Does human attention follow the reasoning process?}
First, to analyze the spatiotemporal deployment of human attention in visual reasoning, we conduct a time course analysis by grouping fixations into three temporal bins (\ie~0-1s, 1-2s, and 2-3s) and analyzing both the attention and the allocation of attention. We measure the accuracy of attention with AiR-E scores for each fixation map and reasoning step (see \fig~6b-c), and study the allocation of attention by computing the Correlation Coefficient (CC)~\cite{sal_metric} between fixation maps and a center prior baseline created by placing a Gaussian ($\sigma=15$) at the image center~\cite{borji_iccv13}. Our results show that humans start exploring the visual scene (\ie~0-1s) with relatively low attention accuracy because it takes time to understand the visual scene and locate the correct ROIs. Their attention is also more biased towards the central regions at the beginning because of the experimental setting that aligns the initial fixation with the image center, and the advantage of rapidly extracting the gist of the scene \cite{center_bias_adv}, \ie~0.47 CC score for the first second compared to 0.15 CC score for the latter periods. After the initial exploration, human attention shows improved accuracy across all reasoning steps (\ie~1-2s), and particularly focuses on the early-step ROIs. In the final steps (\ie~2-3s), depending on the correctness of the answers, human attention either shifts to the ROIs at later stages (\ie~correct), or becomes less accurate with lowered AiR-E scores (\ie~incorrect). Such observations suggest a high spatiotemporal alignment between human attention and the sequence of reasoning operations.

\textbf{Does machine attention follow the reasoning process?}
Similarly, we evaluate multi-glimpse machine attention mechanisms. We compare the stacked attention from SAN~\cite{san}, compositional attention from MAC~\cite{mac} and the multi-head attention~\cite{mcb,mfb}, which all adopt object-based attention. \fig~\ref{fig:step_comparison}d-f shows that multi-glimpse attention mechanisms do not evolve with the reasoning process. Stacked attention's first glimpse already attends to the ROIs at the 4th step, and the other glimpses contribute little to the attention accuracy. Compositional attention and multi-head attention consistently align best with the ROIs at the 3rd or 4th step, and ignore those at the early steps.

The spatiotemporal correlations indicate that following the correct order of reasoning operations is important for humans to attend and answer correctly. In contrast, models tend to directly attend to the final ROIs, instead of shifting their attention progressively.

\subsection{Do Machines and Humans with Diverse Answers Look at Input Images Differently?} \label{diverse_att}
Our previous analyses show various degrees of alignment between the attention, the task outcome, and the intermediate decision-making process. The results motivate us to further study the correlation between attention and task performance, and how different attention patterns lead to diverse answers. In this subsection, we conduct pairwise comparisons between humans or VQA models, and organize the questions into different groups based on the correctness of the two answers. 

Specifically, for each pair of models/humans, questions in the AiR-D dataset can fall into three distinct groups: questions where both humans/models answer them correctly (Correct) or incorrectly (Incorrect), or those where only one human/model answers correctly (Inter).

For the comparison of human attention, we follow \cite{iqva} and measure the alignment between gaze sequences using the edit distance on real sequence (EDR) \cite{edr}, and use the AUC-Judd~\cite{bylinskii2015saliency} to measure the inter-subject consistency in gaze distributions. Lower EDR and higher AUC-Judd measures suggest more consistent attention. Our experiments suggest certain agreements between the correct and incorrect human attention (\ie~EDR=0.641, AUC-Judd=0.872). The inter-subject agreement within the correct attention group is high (\ie~EDR=0.592, AUC-Judd=0.895) while that with the incorrect attention group is relatively low (\ie~EDR=0.635, AUC-Judd=0.864). 

For the comparison of machine attention, for each question, we measure the Spearman's Rank correlation $r$ between attentions corresponding to the two models. Table \ref{pairwise_att_alignment} reports the results for machine attention. We choose UpDown and MUTAN with soft attention (S) and Transformer attention (T) as the backbone models, as they have comparable VQA accuracy on the GQA validation set. All models adopt the object-based attention.

\begin{table}
\begin{center}
\begin{tabular}{cccc}
\toprule
 & Inter & Correct & Incorrect \\
\midrule
UpDown (S) - MUTAN (S) & 0.569 & 0.610 & 0.698 \\
UpDown (T) - MUTAN (T) & 0.308 & 0.460 & 0.546\\
\midrule
UpDown (S) - UpDown (T) & 0.440 & 0.575 & 0.634 \\
MUTAN (S) - MUTAN (T) & 0.397 & 0.444 & 0.528 \\
\midrule
UpDown (S) - MUTAN (T) & 0.440 & 0.475 & 0.556 \\
UpDown (T) - MUTAN (S) & 0.422 & 0.523 & 0.602 \\
\bottomrule
\end{tabular}
\end{center}
\caption{Spearman's Rank Correlation between machine attention mechanisms for different answers. For each group separated by the horizontal lines, from top to bottom are results on different VQA backbones but the same type of attention, the same backbone but with different types of attention, and different backbones and attention mechanisms.}
\label{pairwise_att_alignment}
\end{table}

\begin{table*}
\begin{center}
\begin{tabular}{ccccccc}
\toprule
&  \multicolumn{2}{c}{UpDown \cite{updown}} & \multicolumn{2}{c}{MUTAN \cite{mutan}} & \multicolumn{2}{c}{BAN \cite{ban}}\\
\cmidrule{2-7}
  & dev  & standard & dev  & standard & dev  & standard   \\
\midrule
w/o Supervision & 51.31 & 52.31 & 50.78 & 51.16 & 50.14 & 50.38 \\
PAAN \cite{paan} & 48.03 & 48.92 & 46.40 & 47.22 & n/a & n/a \\ HAN \cite{han} & 49.96 & 50.58 & 48.76 & 48.99 & n/a & n/a\\
ASM \cite{mining} & 52.96 & 53.57 & 51.46 & 52.36 & n/a & n/a\\ 
AiR-M & \textbf{53.46} & \textbf{54.10} & \textbf{51.81} & \textbf{52.42} & \textbf{53.36} & \textbf{54.15} \\
\bottomrule
\end{tabular}
\end{center}
\caption{Comparative results on GQA test sets (test-dev and test-standard). All the compared results are from single models trained on the balanced training set of GQA.}
\label{model}
\end{table*}

\begin{figure*}
\centering
\includegraphics[width=0.7\linewidth]{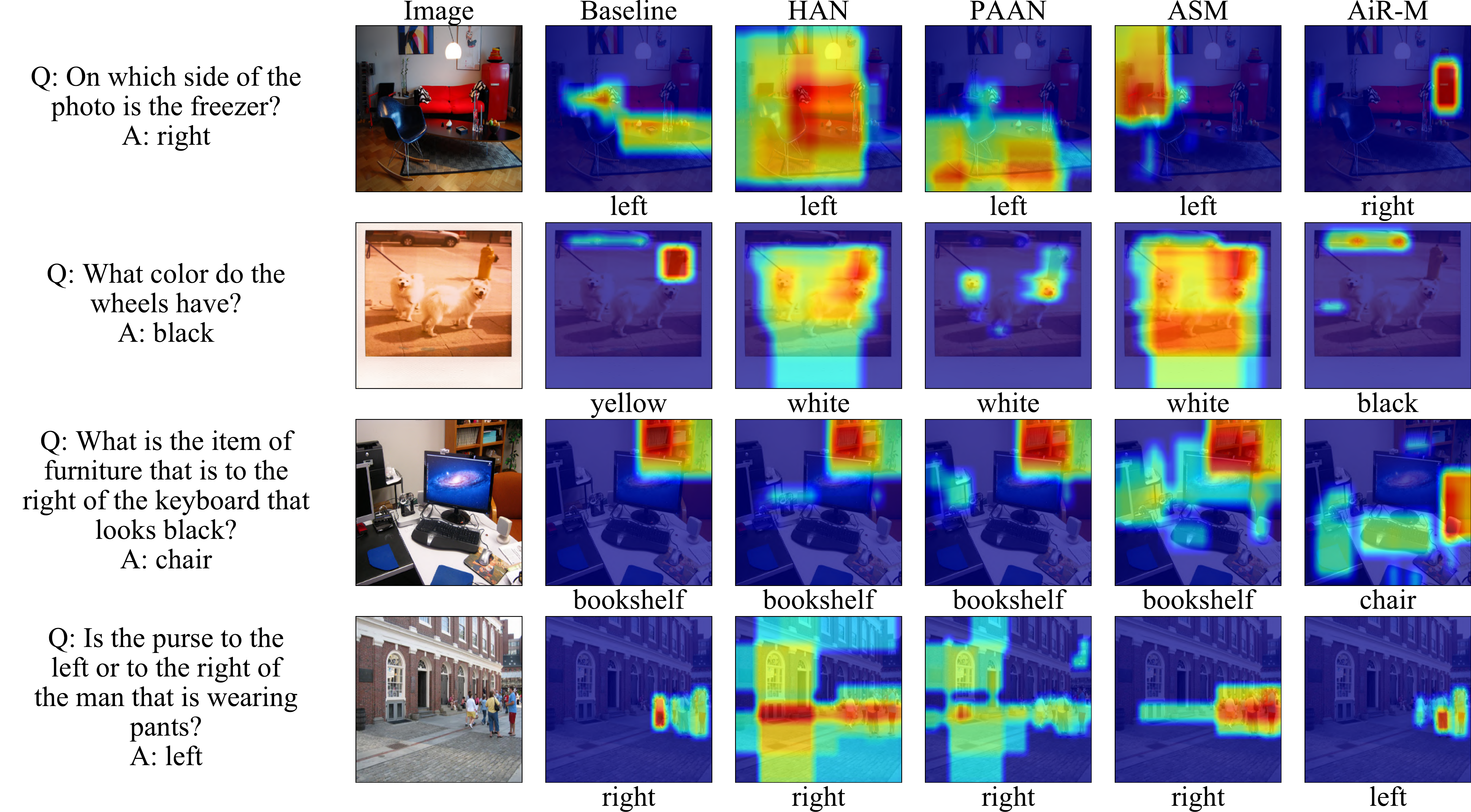}
\caption{Qualitative comparison between attention supervision methods, where Baseline refers to UpDown~\cite{updown}. For each row, from left to right are the questions and the correct answers, input images, and attention maps learned by different methods. The predicted answers associated with each attention mechanism are shown below its respective attention map.}
\label{fig:model_qualitative}
\end{figure*}

Three observations can be made from the experimental results: 
\begin{enumerate*}
\item[(1)] Both humans and models have higher diversity of attention if their answers are different. Compared to the alignment scores for both the correct attention and the incorrect attention, the inter-correctness alignment scores are consistently lower.
\item[(2)] Humans tend to converge on similar ROIs to answer questions, while machines tend to have more diverse focuses, depending on both the backbone models and attention types. This is validated by the high variance of attention alignment scores across different models being compared. 
\item[(3)] Compared with humans, models are more vulnerable to the most salient distractors, as they have higher alignment scores for incorrect attention.
\end{enumerate*}

The aforementioned observations reveal the visual behaviors of humans and machines when deriving different answers. More importantly, it shows that, unlike humans, models are vulnerable to similar hard-negative distractors when answering a question, suggesting the usefulness of incorporating the negative attention to encourage models to avoid these distractors.

\subsection{Does Progressive Attention Supervision Improve Attention and Task Performance?} \label{att_supervision}
Experiments in Section \ref{att_outcome} and Section \ref{att_process} suggest that attention towards ROIs relevant to the reasoning process contributes to task performance, and furthermore, the order of attention matters. Therefore, we propose to guide models to look at places important to reasoning in a progressive manner. Specifically, we propose to supervise machine attention throughout the reasoning process by jointly optimizing attention, reasoning operations, and task performance (\ie~AiR-M, Section \ref{sup_method}). Here we investigate the effectiveness of the AiR-M supervision method on three VQA models, \ie~UpDown~\cite{updown}, MUTAN~\cite{mutan}, and BAN~\cite{ban}. We compare AiR-M with a number of state-of-the-art attention supervision methods, including supervision with human-like attention (HAN)~\cite{han}, attention supervision mining (ASM)~\cite{mining} and adversarial learning (PAAN)~\cite{paan}. Note that while the other compared methods are typically limited to supervision on a single attention map, our AiR-M method is generally applicable to various VQA models with single or multiple attention maps (\eg~BAN~\cite{ban}).

\begin{table}
\begin{center}
\begin{tabular}{c c}
\toprule
Method & GQA test-dev\\
\midrule
AiR-M w/o $L_{\Vec{\alpha}}$ & 50.01 \\
AiR-M w/o $L_{\Vec{r}}$ & 50.33 \\
AiR-M Single & 52.84 \\
AiR-M & \textbf{53.46} \\
\bottomrule
\end{tabular}
\caption{Experimental results of AiR-M under different supervision strategies. All reported results are on the GQA \cite{gqa} test-dev set. Bold numbers indicate the best performance.}
\label{ablation_optim}
\end{center}
\end{table} 

\begin{table*}
\begin{center}
\begin{tabular}{lllllllll}
\toprule
Attention &             and &         compare &           filter &              or &           query &           relate &           select &          verify \\
\midrule
Human   &  2.197 &    2.669 &   2.810 & 2.429 &  3.951 &   3.516 &   2.913 &   3.629\\
\midrule
AiR-M       & \textbf{2.396} &    \textbf{2.553} &   \textbf{2.383} & \textbf{2.380} &  3.340 &   \textbf{2.862} &   \textbf{2.611} &   \textbf{4.052} \\
Baseline \cite{updown} & 1.859 &    1.375 &   1.717 & 2.271 &  \textbf{3.651} &   2.448 &   1.796 &   2.719 \\
ASM    & 1.415 &    1.334 &   1.443 & 1.752 &  2.447 &   1.884 &   1.584 &   2.265 \\
HAN       & 0.581 &    0.428 &   0.468 & 0.607 &  1.576 &   0.923 &   0.638 &   0.680 \\
PAAN      & 1.017 &    0.872 &   1.039 & 1.181 &  2.656 &   1.592 &   1.138 &   1.221 \\
\bottomrule
\end{tabular}
\end{center}
\caption{AiR-E scores of the supervised attention mechanisms.}
\label{corr_supervision}
\end{table*} 

As shown in \tab~\ref{model}, the proposed AiR-M method significantly improves the performance of all baselines and consistently outperforms the other attention supervision methods. Two of the compared methods, HAN and PAAN, fail to improve the performance of object-based attention. Supervising attention with knowledge from objects mined from language, ASM~\cite{mining} can consistently improve the performance of models. However, without considering the intermediate steps of reasoning, it is not as effective as the proposed method. In addition to the enhanced VQA performance, our method also predicts the reasoning steps with high accuracy ($96.2\%$ validation accuracy on reasoning step prediction). It shows that our method can accurately capture the correct reasoning process and learn reasoning-aware attention to improve the performance of visual reasoning. 

\fig~\ref{fig:model_qualitative} shows the qualitative comparison between supervision methods. As the previous supervision methods (\ie~HAN, PAAN and ASM) are optimized to simultaneously capture all important regions, their attention outputs tend to spread over multiple ROIs, in which some are less relevant. On the contrary, by progressively supervising the attention throughout the reasoning process, our proposed AiR-M learns focused attention towards the most relevant ROIs (\ie~freezer, wheel, chair, purse). Moreover, unlike reasoning-agnostic methods that commonly ignore ROIs for intermediate decision-making steps (\ie~keyboard, man), our method can capture diverse ROIs with regard to the entire reasoning process.

Our method jointly and progressively optimizes attention and reasoning operations. To further demonstrate its advantages, we compare it with three alternatives: two models trained with either attention ground truth or reasoning operations (AiR-M w/o $L_{\Vec{r}}$ and AiR-M w/o $L_{\Vec{\alpha}}$), and a single-glimpse model jointly optimized on both types of ground truth (AiR-M Single, the attention is supervised with ground truth aggregated across all reasoning steps). Three observations can be drawn from the results in Table~\ref{ablation_optim}: (1) Due to the lack of knowledge about the correlation between attention and reasoning process, individually optimizing the reasoning semantics or fine-grained grounding fails to improve the performance; (2) Jointly optimizing attention and the reasoning process with the same types of ground truth (AiR-M Single) leads to significant improvements, demonstrating the need of learning reasoning-aware attention; and (3) Compared to single-step attention optimization (AiR-M Single), AiR-M with multi-step progressive supervision can learn more fine-grained attention specific to each reasoning step, resulting in better performance.

\begin{figure}
\centering
\includegraphics[width=0.5\linewidth]{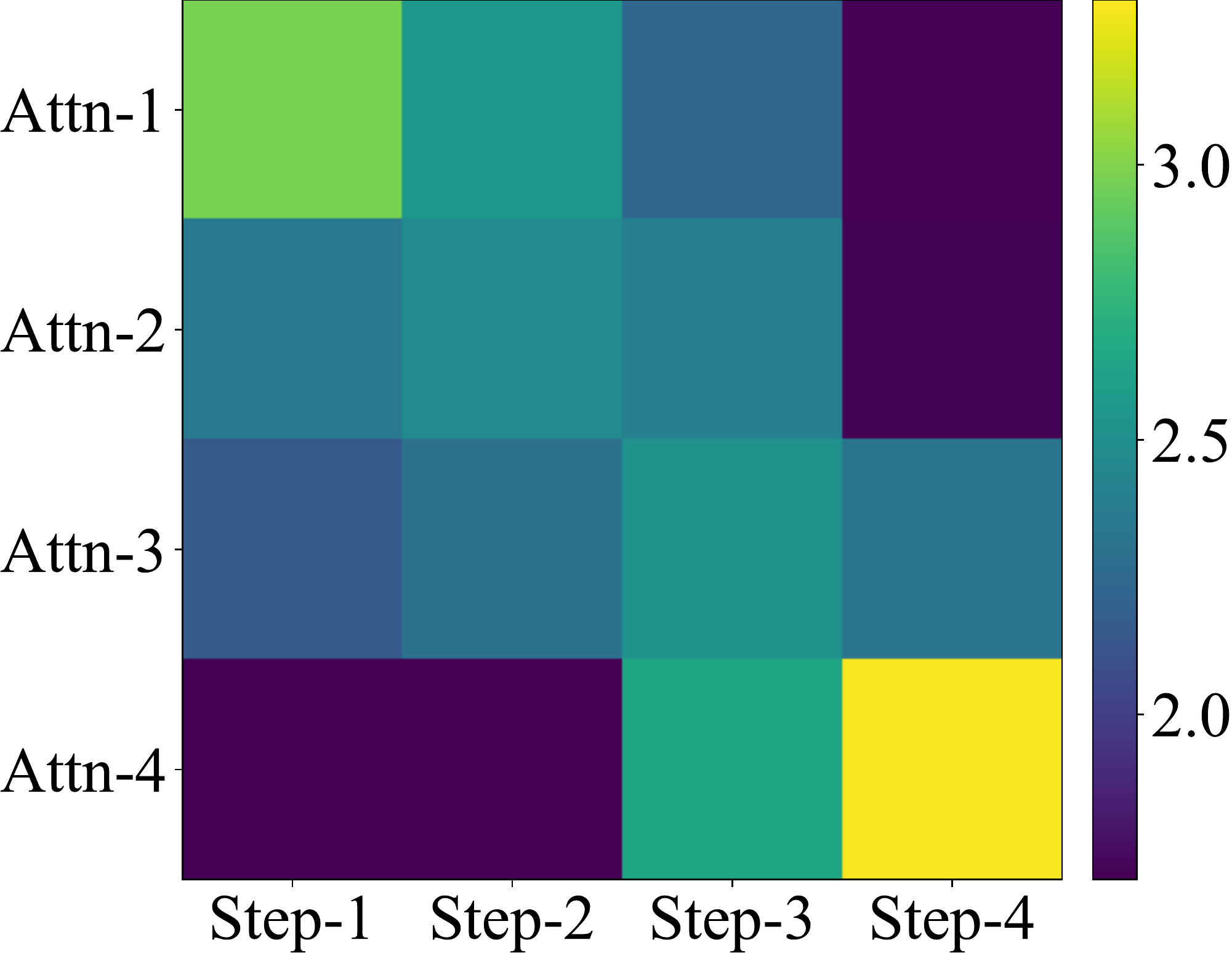}
\caption{Alignment between the attention and reasoning process supervised with the AiR-M method.}
\label{fig:ours}
\end{figure}

To further demonstrate the impact of our AiR-M method on the attention accuracy, \tab~\ref{corr_supervision} reports the AiR-E scores across different operations. It shows that the AiR-M supervision method significantly improves attention accuracy (attention aggregated across different steps), especially on those typically positioned in early steps (\eg~\textit{select}, \textit{compare}). In addition, the AiR-M supervision method also aligns the multi-glimpse attention better according to their chronological order in the reasoning process (see \fig~\ref{fig:ours} and the supplementary video), showing progressive improvement of attention throughout the entire process.  

\begin{figure*}
\centering
\includegraphics[width=0.75\linewidth]{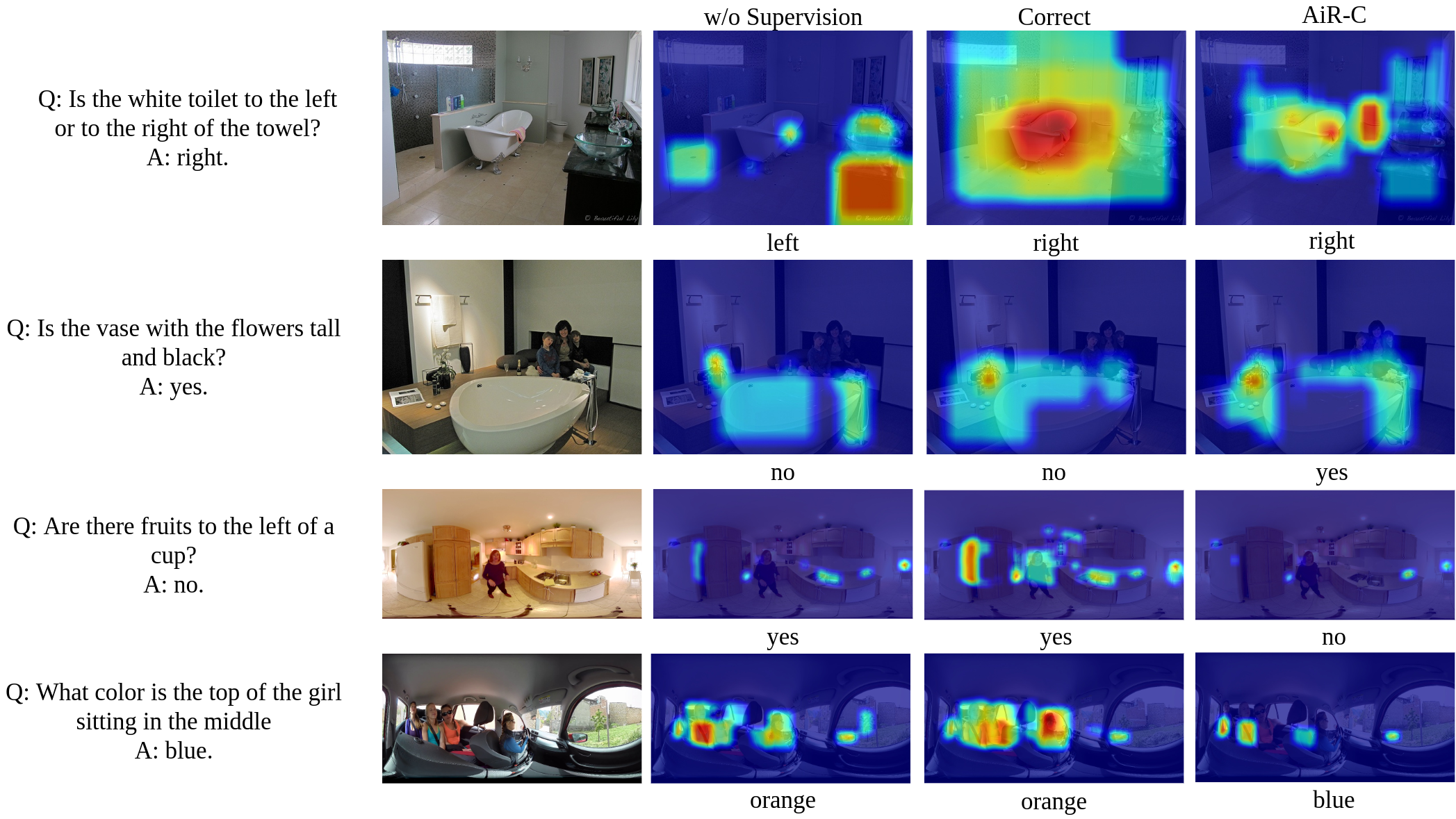}
   \caption{Qualitative results for attention supervision with incorrect attention. For sample show in each row, from left to right are input image with ground truth question and answer, model attention learned without supervision, model attention learned with our correct attention \cite{mining}, model attention learned with our AiR-C method. The model predicted answers are shown at the bottom.}
\label{neg_qual}
\end{figure*}

\subsection{Does Incorporating the Incorrect Attention Benefit Attention Learning?} \label{neg_sup_sec}
Analyses in Section \ref{diverse_att} show that VQA models tend to predict wrong answers because of hard-negative distractors. In this subsection, we investigate if explicitly supervising models with both correct and incorrect attention (AiR-C) can help them avoid such hard-negative distractors and improve answer accuracy. 
We utilize UpDown \cite{updown} as our backbone model, and conduct experiments on the GQA dataset by supervising the machine attention with the correct and incorrect attention mined from the annotations. 
Further, to demonstrate its generalizability, we experiment our method on the IQVA \cite{iqva} dataset with eye-tracking data on 360$^{\circ}$ videos. Following~\cite{360_vqa}, we decompose the 360$^{\circ}$ visual frames into perspective cubemaps, and apply the UpDown~\cite{updown} backbone on each cubemap. Features from different cubemaps and time steps are combined with trainable attention to derive the final answer. The new model (UpDown-360) is first pre-trained on the GQA dataset, and then fine-tuned on the training set of IQVA.

\begin{table}
\begin{center}
\begin{tabular}{cccc}
\toprule
& \multicolumn{2}{c}{UpDown \cite{updown}}  & UpDown-360 \\
\cmidrule{2-4}
& GQA-dev  & GQA-standard & IQVA \\
\midrule
w/o Supervision & 51.31 & 52.31 & 39.73 \\
Correct & 52.96 & 53.57 & 40.55 \\
AiR-C & \textbf{53.74} & \textbf{53.85} & \textbf{41.10} \\
\bottomrule
\end{tabular}
\end{center}
\caption{VQA accuracy on the GQA test sets (test-dev and test-standard) and IQVA test set. AiR-C denotes our full method incorporating both the correct attention and the incorrect attention.}
\label{neg_sup_res}
\end{table}

Table \ref{neg_sup_res} shows quantitative results of AiR-C compared with two alternatives (\ie~w/o Supervision and supervision with Correct answers). By incorporating both correct and incorrect attention, our AiR-C method can outperform these two counterparts, suggesting that avoiding hard-negative distractors is complementary to the supervision from the correct attention. The improvement brought by the incorrect attention supervision is consistent across different datasets. In addition to the increase in VQA performance, the improvement of attention accuracy (\ie~AiR-E) is also significant. Compared to model supervised by only the correct attention (AiR-E=$1.74$), our AiR-C method can alleviate the distraction from visually salient yet question-irrelevant regions and achieve much higher attention accuracy (AiR-E=$2.02$).

Qualitatively, Fig.~\ref{neg_qual} shows that supervising the models (\ie~UpDown and UpDown-360) with incorrect attention leads to focused attention on the correct ROIs. In the 1st and 2nd examples (perspective images from GQA), other models (\ie~w/o Supervision and Correct) either are distracted by the dominant objects (\ie~cabinet and bathtub), or fail to focus on the correct ROIs (\ie~toilet, towel, and vase), while our AiR-C method helps the model avoid these distractors and focus on the most relevant ROIs to generate correct answers. In the 3rd and 4th examples (\ie~360$^{\circ}$ video frames from IQVA), without knowledge of the visual distractors, other models do not have a clear focus due to the complexity of scenes, while our method develops focused attention.

Theses results demonstrate the effectiveness of incorporating knowledge from hard-negative distractors and suggest the generalizability of the proposed AiR-C method.

\subsection{Do Attention Accuracy and Reasoning Performance Agree?} \label{att_ans_agreement}

Our analyses in the previous sections demonstrate the positive correlation between attention accuracy and reasoning performance (Section \ref{att_outcome}), and show that learning more accurate attention leads to a considerable improvement in reasoning performance (Section \ref{att_supervision} and Section \ref{neg_sup_sec}). To further analyze the impacts of attention on visual reasoning, we conduct an ablation study by replacing the model attention outputs with two extreme types of attention: random attention and ground-truth attention. Therefore, the reasoning accuracy based on the random attention can be seen as the performance lower bound, while the reasoning accuracy based on the ground-truth attention can be seen as the performance upper bound. Specifically, to evaluate the performance lower bound, we replace the attention computed from the pre-trained UpDown \cite{updown} model with randomly sampled attention maps following a uniform distribution. Similarly, by replacing the model's attention with the ground-truth attention, we can evaluate its performance upper bound. With this experiment, we find that the random attention leads to a significant drop in the answer accuracy (-7.39$\%$) over the pre-trained baseline, while the ground-truth attention improves the answer accuracy by a large margin (+8.00$\%$). These performance bounds suggest the significant role of attention in visual reasoning.

\begin{figure*}
\centering
\includegraphics[width=0.7\linewidth]{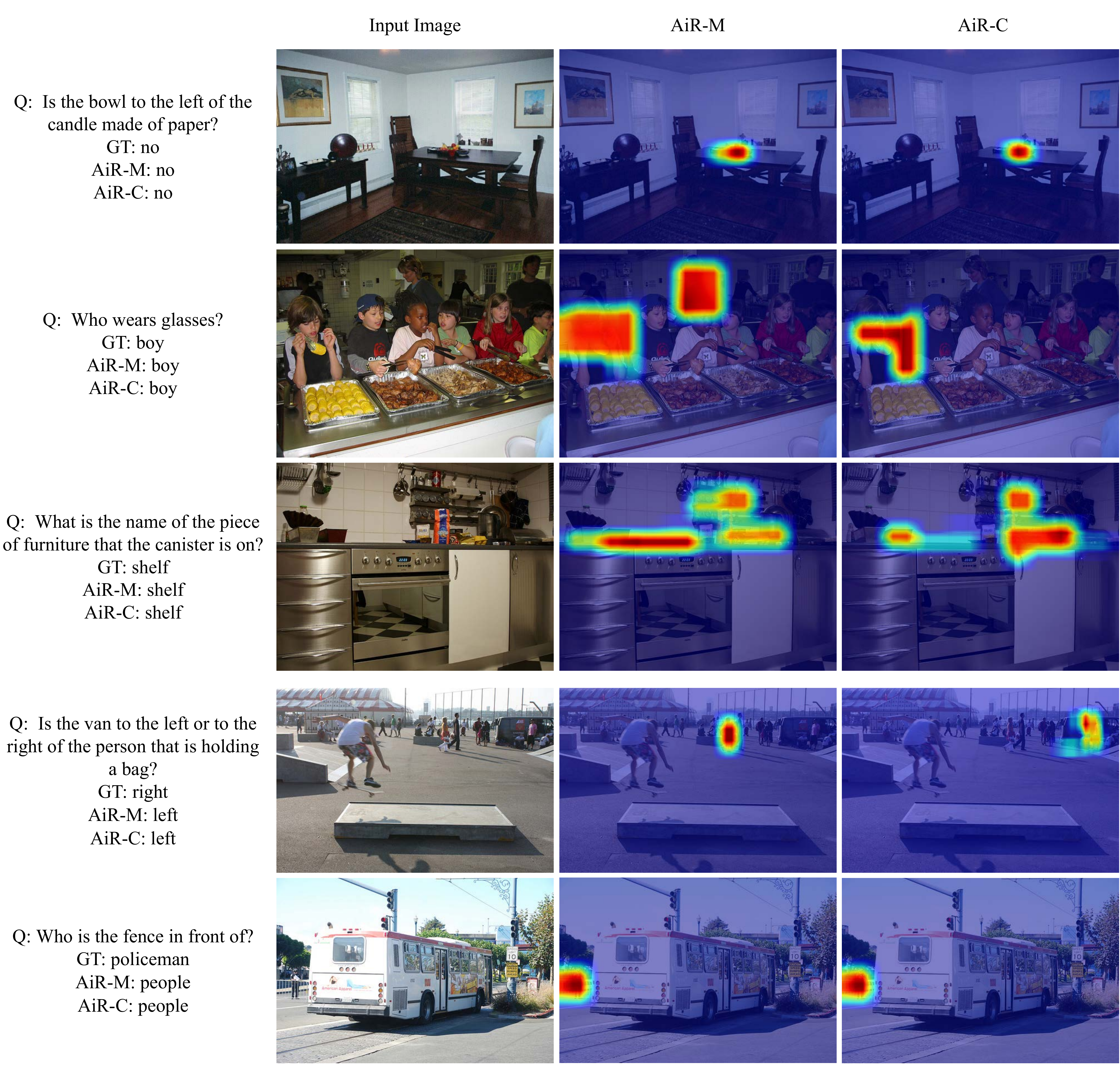}
\caption{Examples for studying the inconsistency between attention accuracy and reasoning performance. From left to right are questions with ground truth (GT) and predicted answers, input images, and attention maps for the two models.}
\label{fig:failure_case}
\end{figure*}

However, attention is not the only important factor for achieving high reasoning accuracy. For instance, visual recognition is also consequential. To correctly answer a question, even with correct attention, one must recognize the attributes of the attended objects and the relationship between them. As a result, there are cases where attention accuracy does not agree with the reasoning performance. Fig.~\ref{fig:failure_case} shows typical cases where the attention accuracy and reasoning performance are inconsistent. These cases include (1) when the model answers correctly but with wrong attention (\ie~AiR-E $<1$), and (2) when the model answers incorrectly but with reasonable attention (\ie~AiR-E $>2.5$). We use our AiR-M and AiR-C methods for demonstration due to their high attention accuracy.

In some cases, the model answers correctly with incorrect attention, which is resulted from various reasons:
\begin{enumerate*}
    \item[(1)] \textbf{Biased prior distribution of questions and answers}. Language biases are prevalent in VQA datasets due to the imbalanced prior distributions of questions and answers. For example, paper bowls are less common, as shown in the 1st example of Fig. \ref{fig:failure_case}. Models leveraging such biases can predict the correct answers without attending to the correct ROIs.
    \item[(2)] \textbf{Attending to wrong objects that coincidentally relate to the answers}. Many images contain abundant objects that may share similar characteristics. As a result, even with incorrect attention, models can still answer correctly by coincidentally looking at another object that relates to the answer. \Eg,~looking at boys not wearing glasses also leads to the correct answer, as shown in the 2nd example of Fig. \ref{fig:failure_case}.
    \item[(3)] \textbf{Capturing the ROIs without focused attention}. For scenes cluttered with various semantics, models may not focus on the correct ROIs. \Eg,~boxes with bright colors attract more attention than the shelf behind them, as shown in the 3rd example of Fig. \ref{fig:failure_case}. However, since the features of the ROIs can be extracted without strong attention, they can still answer correctly despite the low attention accuracy.
\end{enumerate*}

There are also cases where the model answers incorrectly but with reasonable attention:
\begin{enumerate*}
\item[(1)] \textbf{Missing the ROIs directly related to the answers}. Many questions in our dataset require reasoning over multiple ROIs, even if models focus on most of the ROIs, \Eg,~as shown in the 4th example of Fig.~\ref{fig:failure_case}, AiR-M looks at people with the bag but not the van, while AiR-C looks at the van but not the people. They both answer incorrectly due to the failure of capturing both ROIs.
\item[(2)] \textbf{Failing to recognize the ROIs}. Some of the ROIs could be small and difficult to recognize, so models looking at the correct ROIs can still answer incorrectly. \Eg,~as shown in the 5th example of Fig.~\ref{fig:failure_case}), models fail to describe the policeman due to erroneous recognition.
\end{enumerate*}

In sum, these results suggest that the attention accuracy strongly correlates with the reasoning performance in general, but answer correctness is not completely dependent on the accuracy of attention.

\section{Conclusion}\label{sec:conclusion}
We introduce AiR, a novel framework with a quantitative evaluation metric (AiR-E), two supervision methods (AiR-M and AiR-C), and an eye-tracking dataset (AiR-D) for understanding and improving attention in the reasoning context. Our experiments analyze the correlation between attention and task performance in various aspects, and highlight the significant gap between machines and humans on the alignment of attention and reasoning process. With the newly proposed supervision methods, we show that accurate attention deployment can lead to improved task performance, which is related to both the task outcome and the intermediate reasoning steps. We hope that this work will be helpful for the future development of visual attention and reasoning method, and inspire the analysis of model interpretability throughout the reasoning process.

\section*{Acknowledgements}
This work is supported by NSF Grants 1908711 and 1849107.


%
\ifCLASSOPTIONcaptionsoff
  \newpage
\fi



%
%

\bibliographystyle{IEEEtran}
\bibliography{egbib}

\begin{IEEEbiography}[{\includegraphics[width=1in,height=1.25in,clip,keepaspectratio]{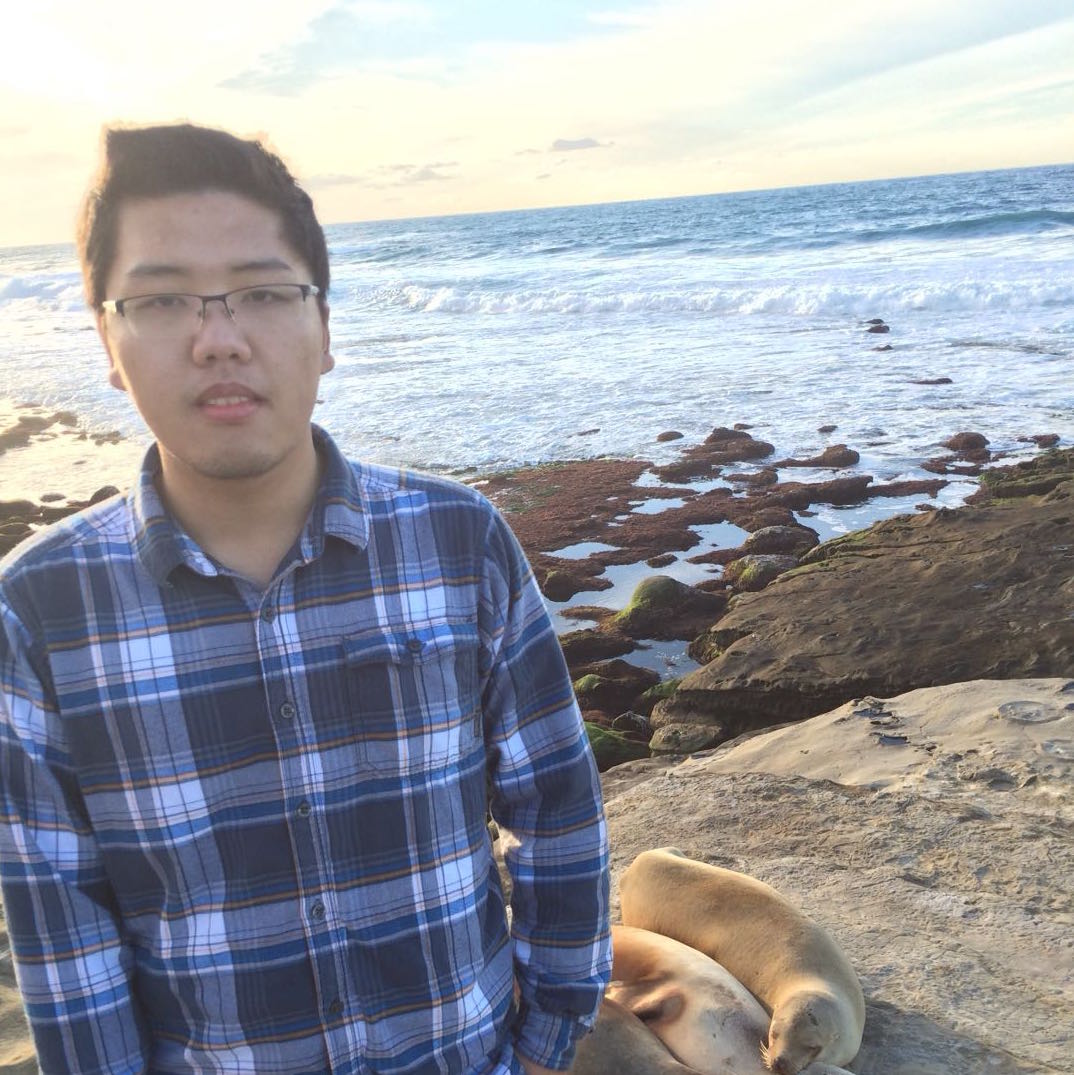}}]{Shi Chen} received the BE degree from the
School of Computer Science, Wuhan University,
Wuhan, China, in 2015, and the MS degree
from the University of Minnesota, Minneapolis, in
2017. He is currently working toward the PhD
degree in the Department of Computer Science,
University of Minnesota. His research interests
include computer vision, vision and language, and
machine learning.
\end{IEEEbiography}

\begin{IEEEbiography}[{\includegraphics[width=1in,height=1.25in,clip,keepaspectratio]{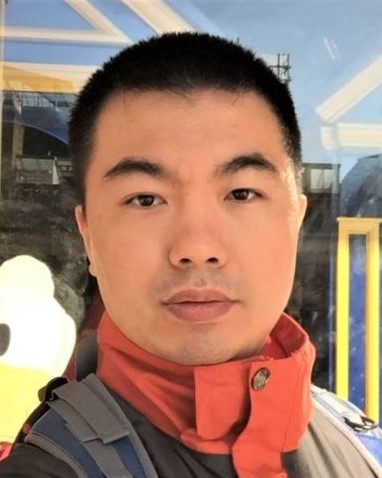}}]{Ming Jiang} is a researcher at the Department of Computer Science and Engineering, University of Minnesota. He obtained his Ph.D. degree in Electrical and Computer Engineering from the National University of Singapore. His M.E and B.E degrees were obtained from Zhejiang University, Hangzhou, China. His research interests are computer vision, cognitive vision, machine learning, psychophysics, neuroscience, and brain-machine interface.
\end{IEEEbiography}

\begin{IEEEbiography}[{\includegraphics[width=1in,height=1.25in,clip,keepaspectratio]{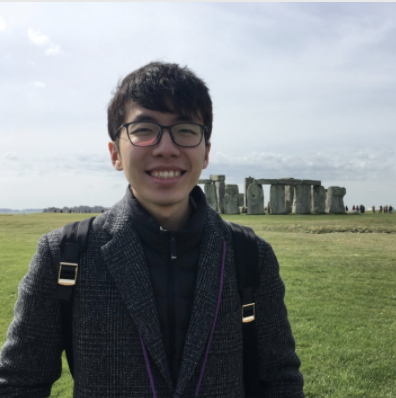}}]{Jinhui Yang} is a Ph.D. student at the Department of Computer Science and Engineering, University of Minnesota. He graduated from Carleton College in 2019 with a BA degree in Computer Science and Statistics. His current research interests include computer vision, interpretable machine learning, and deep neural networks.
\end{IEEEbiography}

\begin{IEEEbiography}[{\includegraphics[width=1in,height=1.25in,clip,keepaspectratio]{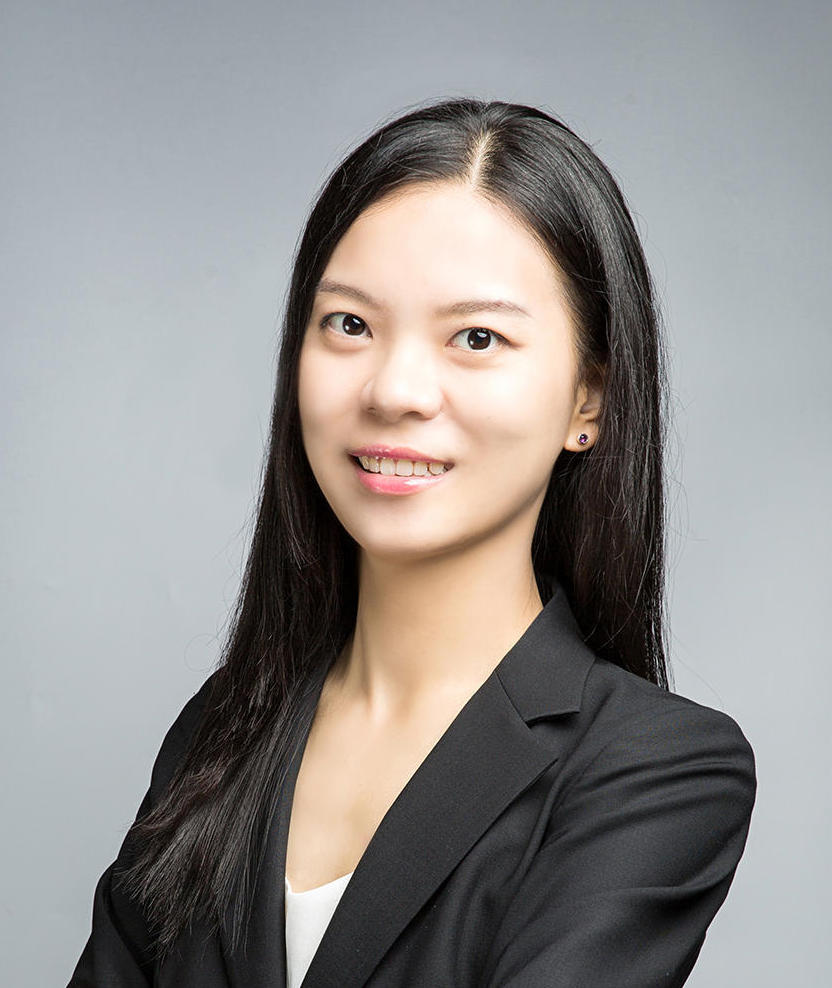}}]{Qi Zhao}
is an associate professor in the Department of Computer Science and Engineering at the University of Minnesota, Twin Cities.
Her main research interests include computer vision, machine learning, cognitive neuroscience,
and healthcare. She received her Ph.D.
in computer engineering from the University of
California, Santa Cruz in 2009. She was a postdoctoral researcher in the Computation \& Neural
Systems, and Division of Biology at the California Institute of Technology from 2009 to 2011.
Before joining the University of Minnesota, Qi was an assistant professor in the Department of Electrical and Computer Engineering and the
Department of Ophthalmology at the National University of Singapore.
She has published more than 100 journal and conference papers in
computer vision, machine learning, and cognitive neuroscience venues,
and edited a book with Springer, titled Computational and Cognitive
Neuroscience of Vision, that provides a systematic and comprehensive
overview of vision from various perspectives.
Qi serves as an associate editor at IEEE Transactions on Neural Networks and Learning Systems (TNNLS) and IEEE Transactions on Multimedia (TMM), as a program chair at IEEE Winter Conference on Applications of Computer Vision (WACV), and as an organizer and/or area chair at IEEE Conference on Computer Vision and Pattern Recognition (CVPR) and other major venues in computer vision and AI. She is a member of the IEEE since 2004.
\end{IEEEbiography}


\vfill


\end{document}